\documentclass[10pt,twocolumn,letterpaper]{article}

\usepackage{times}
\usepackage{graphicx}
\usepackage{amsmath}
\usepackage{amssymb}

\usepackage{color}
\usepackage{bm}
\usepackage{caption}
\usepackage{siunitx} 
\usepackage{adjustbox}
\usepackage{booktabs} 
\usepackage{multirow}
\usepackage{listings}
\usepackage{minibox}
\usepackage{xspace}
\usepackage{threeparttable}
\usepackage{comment}
\usepackage{dcolumn}

\makeatletter
\DeclareRobustCommand\onedot{\futurelet\@let@token\@onedot}
\def\@onedot{\ifx\@let@token.\else.\null\fi\xspace}
\def\eg{\emph{e.g}\onedot} 
\def\ie{\emph{i.e}\onedot}

\def\etal{\emph{et al}\onedot}

\makeatother

\usepackage[T1]{fontenc}
\usepackage[font=small,labelfont=bf,tableposition=top]{caption}
\DeclareCaptionLabelFormat{andtable}{#1~#2  \&  \tablename~\thetable}
\newcommand{\tref}[1]{Table~\ref{#1}}

\newcommand{\eref}[1]{Equation~(\ref{#1})}

\newcommand{\fref}[1]{Figure~\ref{#1}}

\newcommand{\sref}[1]{Section~\ref{#1}}

\newcommand{\rot}{\mathbf{R}}
\newcommand{\trans}{\mathbf{t}}
\newcommand{\extrinsics}{\left[~\rot~|~\trans~\right]}

\newcommand{\vuh}{\tilde{\mathbf{u}}}
\newcommand{\vph}{\tilde{\mathbf{p}}}

\usepackage[pagenumbers]{cvpr} 

\definecolor{cvprblue}{rgb}{0.21,0.49,0.74}
\usepackage[pagebackref,breaklinks,colorlinks,allcolors=cvprblue]{hyperref}

\def\paperID{00000} 
\def\confName{CVPR}
\def\confYear{2026}

\title{Panoramic Distortion-Aware Tokenization for Person Detection and Localization in Overhead Fisheye Images}

\author{
Nobuhiko Wakai$^1$ \quad\quad Satoshi Sato$^1$ \quad\quad Yasunori Ishii$^1$ \quad\quad Takayoshi Yamashita$^2$\\
$^1$ Panasonic Holdings Corporation\quad\quad $^2$ Chubu University\\
{\tt\small \{wakai.nobuhiko,sato.satoshi,ishii.yasunori\}@jp.panasonic.com} \quad\quad {\tt\small takayoshi@isc.chubu.ac.jp}
}

\begin{document}
\maketitle

\begin{abstract}
Person detection in overhead fisheye images is challenging due to person rotation and small persons. Prior work has mainly addressed person rotation, leaving the small-person problem underexplored. We remap fisheye images to equirectangular panoramas to handle rotation and exploit panoramic geometry to handle small persons more effectively. Conventional detection methods tend to favor larger persons because they dominate the attention maps, causing smaller persons to be missed. In hemispherical equirectangular panoramas, we find that apparent person height decreases approximately linearly with the vertical angle near the top of the image. Using this finding, we introduce panoramic distortion-aware tokenization to enhance the detection of small persons. This tokenization procedure divides panoramic features using self-similar figures that enable the determination of optimal divisions without gaps, and we leverage the maximum significance values in each tile of the token groups to preserve the significance areas of smaller persons. We propose a transformer-based person detection and localization method that combines panoramic-image remapping and the tokenization procedure. Extensive experiments demonstrated that our method outperforms conventional methods on large-scale datasets.
\end{abstract}


\section{Introduction}
\begin{figure}[t]
\centering
\includegraphics[width=0.98\hsize]{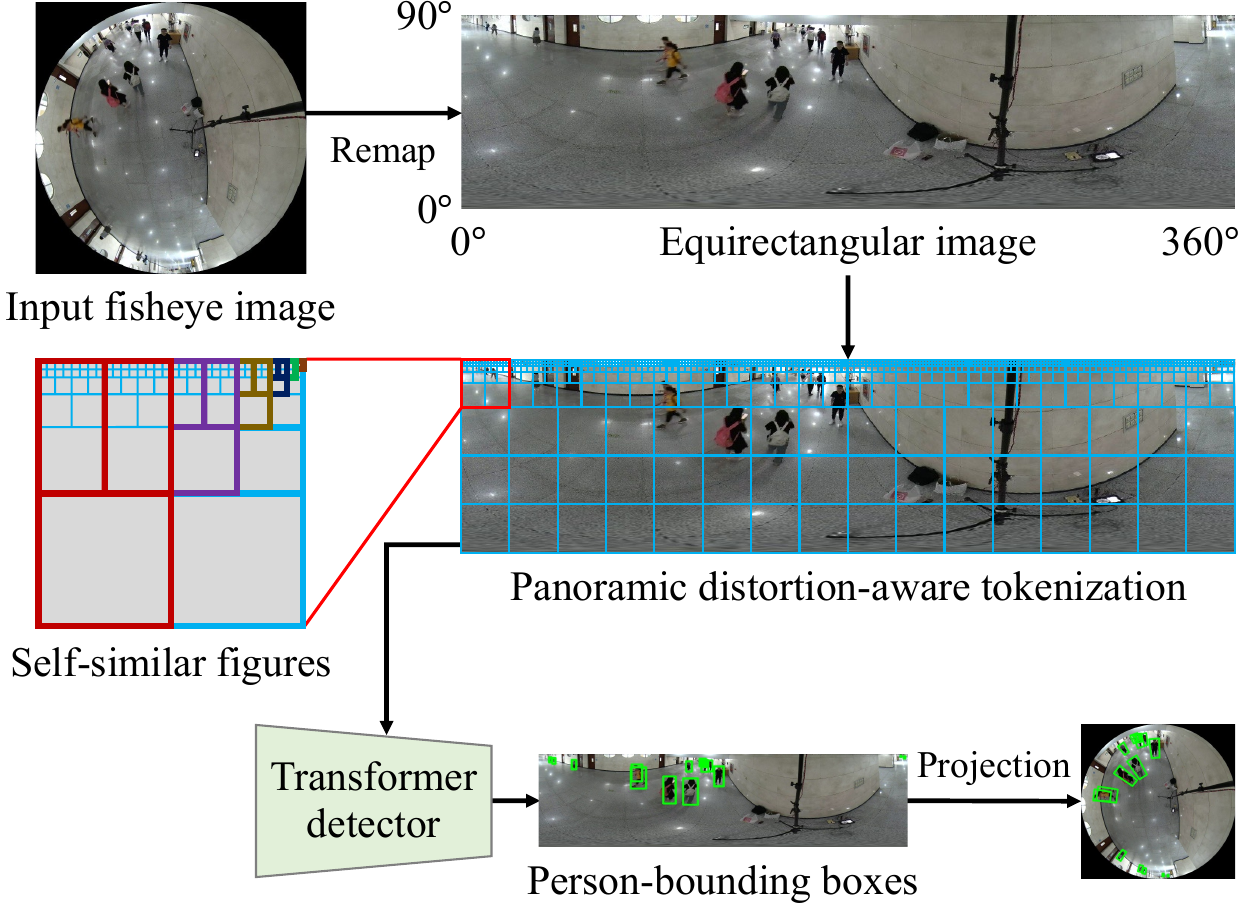}
\caption{Our method converts an overhead fisheye image into an equirectangular image by remapping. To leverage the significance areas of smaller persons, a distortion-aware tokenization process based on a self-similar figure using a unit figure, which is colored dark red and consists of a square and two rectangles. Our transformer-based detector uses the resulting tokens to detect bounding boxes within the panoramic image, and then the boxes project to the fisheye image. The input image is taken from~\cite{Yang_ICCV_2023}.}
\label{fig-concept}
\end{figure}
\begin{table*}[t]
\caption{Comparison of the features of conventional methods with those of our proposed method}
\label{table-comparison-of-related-methods}
\centering
\scalebox{0.72}{
\begin{tabular}{ccccccc}
\hline\noalign{\smallskip}
\multicolumn{2}{c}{\multirow{2}{*}{Method}} & \multirow{2}{*}{Detector~$^1$} & \multirow{2}{*}{Training on overhead fisheye images} & \multirow{2}{*}{Bounding box} & \multicolumn{2}{c}{Core idea to address}\\
\cmidrule(lr){6-7}
~ & ~ & ~ & ~ & ~ & ~Person rotation~ & ~~~Small person~~~ \\
\noalign{\smallskip}
\hline
\noalign{\smallskip}
\multicolumn{2}{c}{CNN-based} \\
Seidel~\etal{}~\cite{Seidel_VISAPP_2019} & VISAPP'19 & YOLOv2~\cite{Redmon_CVPR_2017} & ~ & ~~~~Orthogonal rectangle~~~~ & \checkmark & ~ \\
Li~\etal{}~\cite{Li_AVSS_2019} & AVSS'19 & YOLOv3~\cite{Redmon_arXiv_2018} & ~ & Rotated rectangle & \checkmark & ~ \\
Tamura~\etal{}~\cite{Tamura_WACV_2019} & WACV'19 & YOLOv2~\cite{Redmon_CVPR_2017} & \checkmark & Rotated rectangle & \checkmark & ~ \\
RAPiD~\cite{Duan_CVPRW_2020} & CVPRW'20 & FPN~\cite{Lin_CVPR_2017} & \checkmark & Rotated rectangle & \checkmark & ~ \\
OARPD~\cite{Qiao_MTA_2024} & MTA'24 & YOLOv8m~\cite{Varghese_ADICS_2024} & \checkmark & Rotated rectangle & \checkmark & ~ \\
\hline\noalign{\smallskip}
\multicolumn{2}{c}{Transformer-based} \\
Yang~\etal{}~\cite{Yang_ICCV_2023} & ICCV'23 & ~~~DAB-DETR~\cite{Liu_ICLR_2022}~~~ & \checkmark & Rotated rectangle & \checkmark & ~ \\
\multicolumn{2}{c}{Ours} & DAB-DETR~\cite{Liu_ICLR_2022} & \checkmark & Rotated trapezoid & ~~~~~~\checkmark~~~~~~ & ~~~~~~\checkmark~~~~~~ \\
\hline\noalign{\smallskip}
\multicolumn{7}{l}{\scalebox{1.00}{~$^1$ Each detector was modified to address overhead fisheye images. FPN represents feature pyramid networks.}} \\
\end{tabular}
}
\end{table*}
Object detection methods are used in a wide range of applications in both the commercial and industrial fields, including visual surveillance~\cite{Fu_CVPR_2021, Lin_TIP_2012, Xie_ICPR_2010, Yuan_CVPR_2024}, pedestrian detection~\cite{Dollar_CVPR_2009, Khan_CVPR_2023, Zhang_PAMI_2018}, and robotics~\cite{Lee_ICRA_2018, Yoshimi_IROS_2006, Zhang_IROS_2024}. In particular, the interest in fisheye image-based person detection has been growing because of its varied applications, which include monitoring of public transportation and infrastructure, worker movement optimization in factories, and player movement analysis for use in sports strategy. Overhead fisheye cameras are suitable for use in these applications because they are omnidirectional and have a larger field of view than conventional narrow-view cameras; \ie, one overhead fisheye camera can substitute for several narrow-view cameras. However, the overhead fisheye images do not match the inputs of the detectors~\cite{Ghiasi_CVPR_2021, Huang_2025_CVPR, Wang_CVPR_2023, Zhang_2025_CVPR, Zong_ICCV_2023}, which require perspective images. This mismatch is derived from person rotation and from the presence of small persons in the images. The person rotation factor leads to individual persons having various shapes and appearances in the fisheye images. Within the near image circle of circumferential fisheye images, smaller people are captured at a long distance from the fisheye cameras. Although person detection methods based on use of narrow-view cameras are well-established, deriving accurate techniques for person detection and localization from overhead fisheye images has remained an open challenge. 

Person detection methods have been proposed that used convolutional neural networks (CNNs)~\cite{Krizhevsky_NIPS_2012} and transformers~\cite{Vaswani_NIPS_2017} in overhead fisheye images. Early CNN-based methods~\cite{Li_AVSS_2019, Seidel_VISAPP_2019} used detectors that were trained on perspective images. To improve the performance of the CNN-based methods~\cite{Duan_CVPRW_2020, Qiao_MTA_2024, Tamura_WACV_2019} and a transformer-based method~\cite{Yang_ICCV_2023}, their training was also conducted using overhead fisheye images. However, it is difficult for these methods to detect persons accurately in overhead fisheye images because of the rotation and small-person problems.

On the basis of the observations above, to realize accurate person detection, we propose a transformer-based method that detects rotated bounding boxes from an overhead fisheye image, as illustrated in \fref{fig-concept}. An input fisheye image is converted into a panoramic image using a remapping method. For the panoramic image, we introduce a tokenization method called panoramic distortion-aware tokenization (PDAT) to balance the significance areas of feature maps regardless of person sizes. This tokenization uses self-similar figures to perform nonuniform divisions while maintaining consistent person-to-tile size ratios, based on our finding that person height depends on the vertical image coordinates of the panoramic image. Finally, we obtain bounding boxes in the fisheye images via projection.

To investigate the effectiveness of these proposed methods, we conducted extensive experiments on three large-scale datasets~\cite{Duan_CVPRW_2020, Tezcan_WACV_2022, Yang_ICCV_2023}. This evaluation demonstrated that our proposed method outperforms both the conventional CNN-based~\cite{Duan_CVPRW_2020, Li_AVSS_2019, Qiao_MTA_2024, Seidel_VISAPP_2019, Tamura_WACV_2019} and transformer-based~\cite{Yang_ICCV_2023} methods. The major contributions of our study are summarized as follows:

\begin{itemize}
\item We propose a transformer-based method for person detection and localization in overhead fisheye images that achieves higher accuracy than conventional methods by balancing significance areas of the feature maps.

\item We introduce the PDAT method with optimal divisions based on self-similar figures to address a person's height using the vertical panoramic-image coordinates.

\item We demonstrate the effectiveness of our proposed method by applying it to the challenging LOAF dataset. The average precision for distant person positions when using our method on the LOAF test set is substantially greater than that of the existing state-of-the-art method by 32.0.
\end{itemize}

\section{Related work}
\label{sec-related-work}
\textbf{Overhead-fisheye-image person detection.}
Person detection techniques in overhead fisheye images have been developed using various datasets~\cite{Blanco_SPIC_2021, Duan_CVPRW_2020, Li_AVSS_2019, Tezcan_WACV_2022, Yang_ICCV_2023}. Existing person detection methods for overhead fisheye images can be classified into two categories: CNN-based and transformer-based methods, as listed in~\tref{table-comparison-of-related-methods}. Seidel~\etal{}~\cite{Seidel_VISAPP_2019} proposed a pioneering CNN-based detection method. The method divides a fisheye image into upright image patches that are undistorted using the given camera parameters. Although these patches are perspective images, the method requires camera calibration before detection. Li~\etal{}~\cite{Li_AVSS_2019} proposed a CNN-based method to improve detection accuracy using focal windows that can extract upright images. Tamura~\etal{}~\cite{Tamura_WACV_2019} proposed a method that uses YOLOv2~\cite{Redmon_CVPR_2017}, which had been trained on perspective images from the COCO dataset~\cite{Lin_ECCV_2014} using rotation augmentation. To train a detector called RAPiD, Duan~\etal{}~\cite{Duan_CVPRW_2020} introduced the angle-aware loss function. Inspired by RAPiD being trained on fisheye images, OARPD~\cite{Qiao_MTA_2024} was proposed with use of the center distance intersection over union approach to address rotated bounding boxes. Yang~\etal{}~\cite{Yang_ICCV_2023} proposed a pioneering transformer-based method with rotation-equivariant training. For a query-based detector, this training strategy involves use of an additional query of a $90^\circ$-rotated image to tackle rotated persons. Both the CNN-based and transformer-based methods described above focused primarily on person rotation. Although the CNN-based and transformer-based methods can detect persons, images containing persons of various sizes cause performance degradation.

\textbf{Small object detection.}
Object detection accuracy tends to decrease for smaller objects. Small objects are often captured at long distances, \eg, in a dataset acquired using drones~\cite{Zhu_arXiv_2018}. Small object detection methods are generally classified into four categories: division-based methods, augmentation-based methods, enlargement-based methods, and spectrum-based methods. The division-based methods involve multi-scale training performed using image pyramids~\cite{Singh_CVPR_2018}, image cropping~\cite{Singh_NIPS_2018}, and feature pyramids~\cite{Gong_WACV_2021}. The augmentation-based methods~\cite{Ghiasi_CVPR_2021, Yu_WACV_2020} were proposed to realize data invariance. The enlargement-based methods enlarge the input images~\cite{Liu_TIP_2025, Unel_CVPRW_2019, Wang_PAMI_2021} and conduct super-resolution processing~\cite{Ji_ICPR_2021, Shermeyer_CVPRW_2019}. The spectrum-based methods enhance the discrimination of objects from a frequency perspective~\cite{Sun_2025_CVPR}. The methods noted above do not use the prior knowledge that smaller persons are only located near the image circles in overhead fisheye images. Therefore, these conventional methods do not work effectively for person detection in overhead fisheye images.

\textbf{Transformer tokenization.}
Dosovitskiy~\etal{}~\cite{Dosovitskiy_ICLR_2021} proposed a pioneering transformer-based method using a tokenization approach that divides an input image into square patches. Numerous tokenization methods have been proposed to date because these square patches lead to high computational costs. Window-based methods focus on the token groups using shifted windows~\cite{Liu_ICCV_2021}, cross-shaped windows~\cite{Dong_CVPR_2022}, dilated windows~\cite{Tu_ECCV_2022, Wang_PAMI_2024}, and scalable sliding windows~\cite{Hassani_CVPR_2023}. In contrast to static window shapes, deformable attention methods~\cite{Xia_CVPR_2022} were proposed to improve flexibility. To address the problem of multiple tokens being generated in unimportant areas, \eg, in the background, DynamicViT~\cite{Rao_NIPS_2021} and BiFormer~\cite{Zhu_CVPR_2023} prune the areas, and SG-Former~\cite{Ren_ICCV_2023} aggregates the areas. Although the methods described above are all well-developed, these methods were designed for use with perspective images. For fisheye image applications, DarSwin~\cite{Athwale_ICCV_2023}, which uses radial patches formed by polar partitioning, was proposed. However, this partitioning approach is not suitable for overhead fisheye images because a small number of the relevant tokens are generated in near image circles.

\textbf{Visual localization.}
Various localization methods have been established to determine either camera poses or person locations from an image. These methods can be classified into three categories: query-based methods, regression-based methods, and geometry-based methods. The query-based methods estimate camera poses using databases composed of geo-tagged images~\cite{Aiger_ICCV_2023, Balntas_ECCV_2018, Hyeon_ICCV_2021, Sarlin_CVPR_2021, Torii_PAMI_2021, YangL_ICCV_2019} and 3D scene structure models~\cite{Germain_CVPR_2021, Sarlin_CVPR_2019}. Before query-based methods can be used, the databases and models must first be built in a time-consuming process. The regression-based methods~\cite{Kendall_ICCV_2015, Walch_ICCV_2017} can estimate camera poses without the need for databases and models; however, their inferences can only be applied to the same scenes that were used for training. The geometry-based methods estimate horizontal distances of person positions based on the bounding boxes~\cite{Yang_ICCV_2023, Zhu_JVCIR_2019} and head points~\cite{DelBlanco_SPIC_2021} in overhead fisheye images. These geometry-based methods estimate person locations from image coordinates using camera parameters.

\begin{figure}[t]
\centering
\includegraphics[width=1.00\hsize]{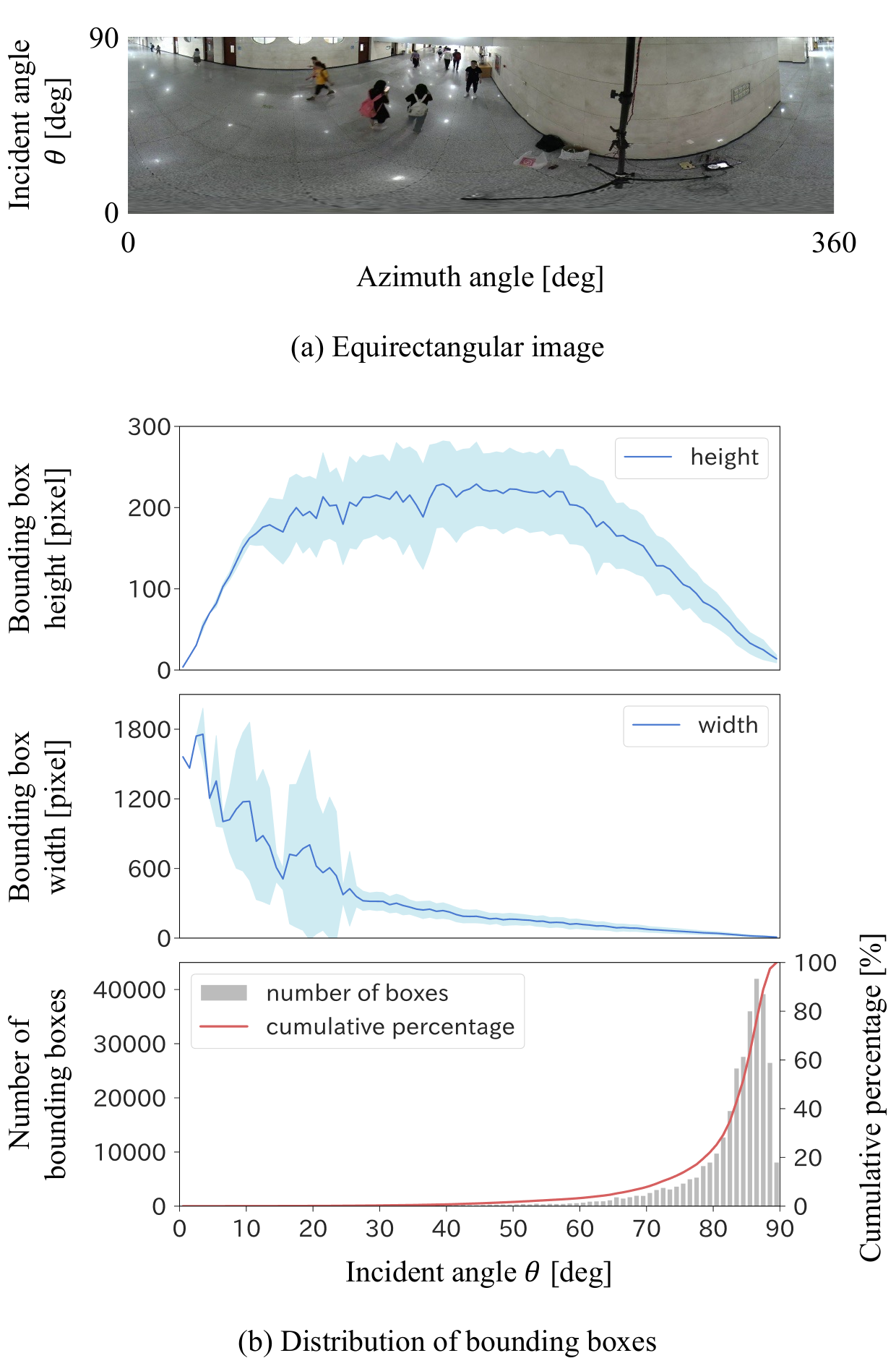}
\caption{Analysis of overhead fisheye images. (a) An equirectangular image. The vertical and horizontal axes show the incident angle (0$^\circ$ is at the bottom of the image) and the azimuth angle, respectively. (b) Distribution of the bounding boxes in equirectangular images on the LOAF training set. This distribution was analyzed using incident angles $\theta$ at 1$^\circ$-intervals based on the center of the bounding boxes. The top and middle panels represent the mean heights and widths of the bounding boxes in equirectangular images with 3072$\times$768 pixels, respectively. The cyan areas indicate $\pm1$ standard deviations in each bin of incident angles. The bottom panel shows the number of bounding boxes.}
\label{fig-bbox-distribution}
\end{figure}
\begin{figure*}[t]
\centering
\includegraphics[width=1.00\hsize]{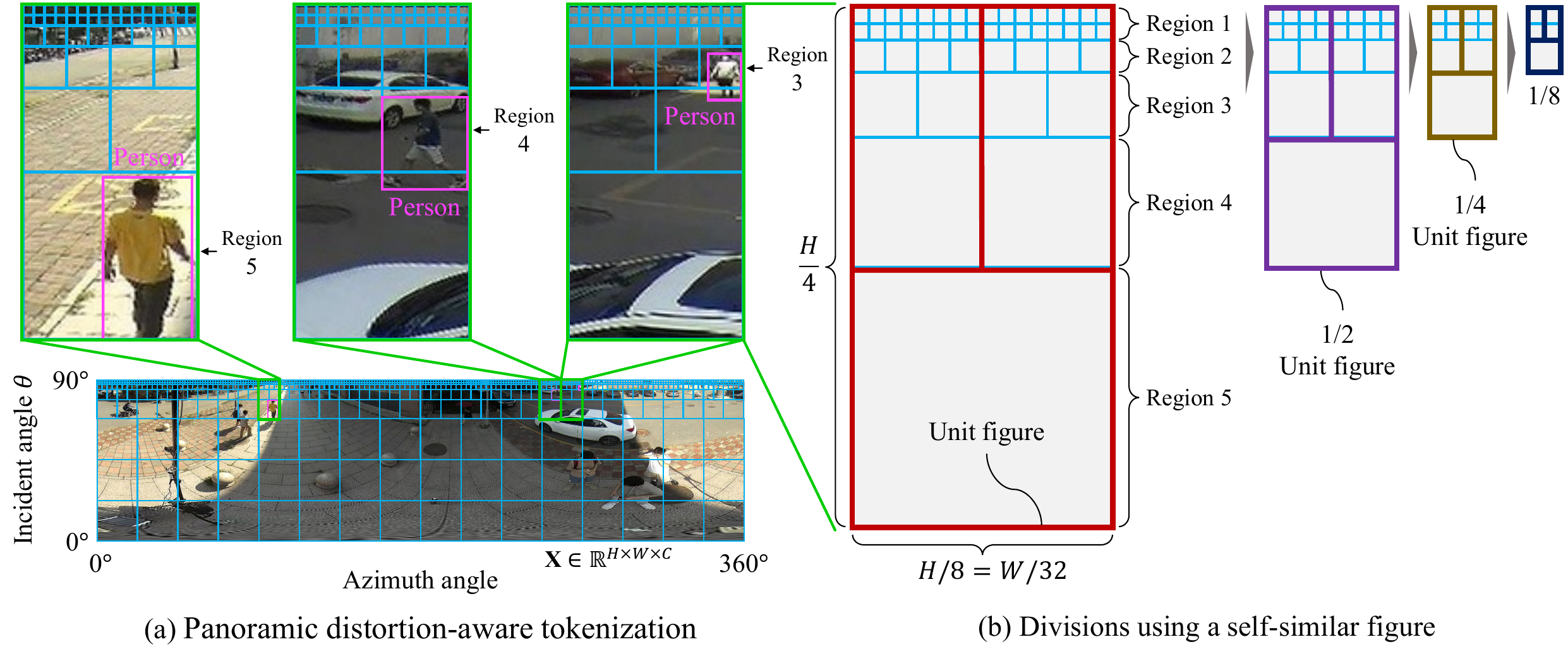}
\caption{Panoramic distortion-aware tokenization. (a) A panoramic feature map is nonuniformly tiled with cyan grids along the incident-angle axis. The tile sizes are adjusted to match the person's height in each region, maintaining similar person-to-tile size ratios across different regions. Enlarged views show person-bounding boxes in regions 3, 4, and 5. (b) A self-similar subdivision recursively partitions the unit figure and is truncated after three steps, yielding five regions ($K=5$) with the size ratios of the unit figure: 1, 1/2, 1/4, and 1/8.}
\label{fig-our-tokenization}
\end{figure*}

\section{Proposed method}
\label{sec-proposed-method}
First, we analyze the bounding boxes in an overhead fisheye image dataset. Second, we present the design of our method. Third, we introduce PDAT. Fourth, we describe our method using PDAT for person detection and localization. Finally, we describe our training and inference phases.

\subsection{Geometric analysis of overhead fisheye images}
\label{sec-geometric-analysis-of-overhead-fisheye-images}
We analyzed the LOAF training set~\cite{Yang_ICCV_2023} to understand the unique characteristics of overhead fisheye images, which differ from perspective images. We remapped overhead fisheye images to equirectangular images, as shown in \fref{fig-bbox-distribution}(a). A downward-looking fisheye camera observes the lower hemisphere, with incident angles $\theta$ ranging from 0$^\circ$ (nadir) to 90$^\circ$ (horizon). \fref{fig-bbox-distribution}(b) shows the distribution of bounding box heights and widths along different incident angles $\theta$. Our analysis revealed two key findings:

\begin{itemize}
\item We found a geometric relationship between equirectangular images and overhead fisheye cameras: the heights and widths of person-bounding boxes decrease approximately linearly when the incident angles $\theta$ are large enough ($>70^\circ$).

\item We observed that 75\% of all person-bounding boxes are concentrated in the range of incident angles from 80$^\circ$ to 90$^\circ$, indicating a strong bias in person distribution.
\end{itemize}
This pattern in the first finding arises from the geometry of overhead fisheye cameras and is fundamentally different from that observed with perspective cameras, where object sizes primarily depend on distance. While the heights at smaller incident angles exhibit an upward-convex trend due to oblique viewing angles, we focus on the linear behavior at large incident angles, where the majority of bounding boxes are located, as indicated in the second finding.

Note that the bias in the second finding has two main causes. First, peripheral portions at large incident angles subtend larger ground areas. Second, due to motion parallax~\cite{Gibson_JEP_1959, Hartley_CUP_2004}, people farther from the camera exhibit smaller apparent image motion and remain in view longer, while nearer people traverse the field of view more quickly.

\subsection{Design of our proposed method}
\label{sec-design-of-our-proposed-method}
We designed our proposed method based on the two key findings, as described in \sref{sec-geometric-analysis-of-overhead-fisheye-images}, because conventional detection methods, which assume uniform object distribution, may not be optimal for overhead fisheye images. The first finding suggests that our method uses the prior knowledge of the linear change in predictable bounding box sizes. The second finding indicates that focusing on bounding boxes with large incident angles is effective in terms of the number of bounding boxes. Conventional detectors allocate computation approximately uniformly across the image and do not explicitly exploit $\theta$-dependent scale or distribution biases. Therefore, we designed our tokenization method to specifically account for these geometric relationships, focusing on the portions with large incident angles. The detailed mathematical derivation of these relationships is provided in the supplementary material.

\subsection{Panoramic distortion-aware tokenization}
\label{sec-panoramic-distortion-aware-tokenization}
On the basis of the design discussed above, we propose PDAT that addresses person detection in overhead fisheye images. The key insight of PDAT is that it maintains consistent person sizes relative to tile sizes across different regions, regardless of their incident angles. To achieve this goal, PDAT employs nonuniform divisions that adapt to the varying sizes of persons at different incident angles, as shown in \fref{fig-our-tokenization}(a). These tiles are designed to fill a panoramic feature map without gaps, ensuring complete coverage for reliable person detection.

As shown in \fref{fig-bbox-distribution}(b), the person's height exhibits an approximately linear decrease at incident angles exceeding 70$^\circ$. This portion corresponds to about the top quarter of the height of the equirectangular images. On the basis of this observation, we align square tiles whose sizes depend on the vertical axis. We found that use of consecutive half-sizes fills the feature maps, as illustrated in \fref{fig-our-tokenization}(a). The top quarter of the image is occupied by performing consecutive half-size tiling, and the other three-quarters consist of tiles of uniform size because the bounding box heights show nonlinear trends in this region with fewer instances.

Each of these square tiles contains $a_k \times a_k$ feature vectors, where $a_k$ represents the number of feature vectors along each side of a tile in the $k$th region, as shown in \fref{fig-our-tokenization}(b). Our tiling is defined on the basis of the self-similar figures when using $K$ regions, with the exception that the top two rows in region 1 have tiles of the same size due to finite divisions. The unit figure consists of a square and two rectangles, and each rectangle can be divided recursively using another set composed of a square and two rectangles, as illustrated in the rightmost part of \fref{fig-our-tokenization}(b). Note that our division method was inspired by the Cantor ternary set~\cite{Smith_LMS_1874}, but our self-similar figure is not a fractal according to Mandelbrot's definition~\cite{Mandelbrot_Science_1967}.

Although the tile sizes vary among regions 3, 4, and 5, as shown in \fref{fig-our-tokenization}(a), the person-bounding boxes appear similar in size relative to their respective tiles. This design maintains similar complexity across all regions, despite the varying absolute sizes of persons in the panoramic image. As illustrated in \fref{fig-our-tokenization}(b), our tokenization method uses recursive divisions to maintain consistent person-to-tile size ratios, particularly in portions with large incident angles. The mathematical formulation of PDAT is given in the supplementary material.

\subsection{Network architecture}
\label{sec-network-architecture}
\begin{figure}[t]
\centering
\includegraphics[width=1.00\hsize]{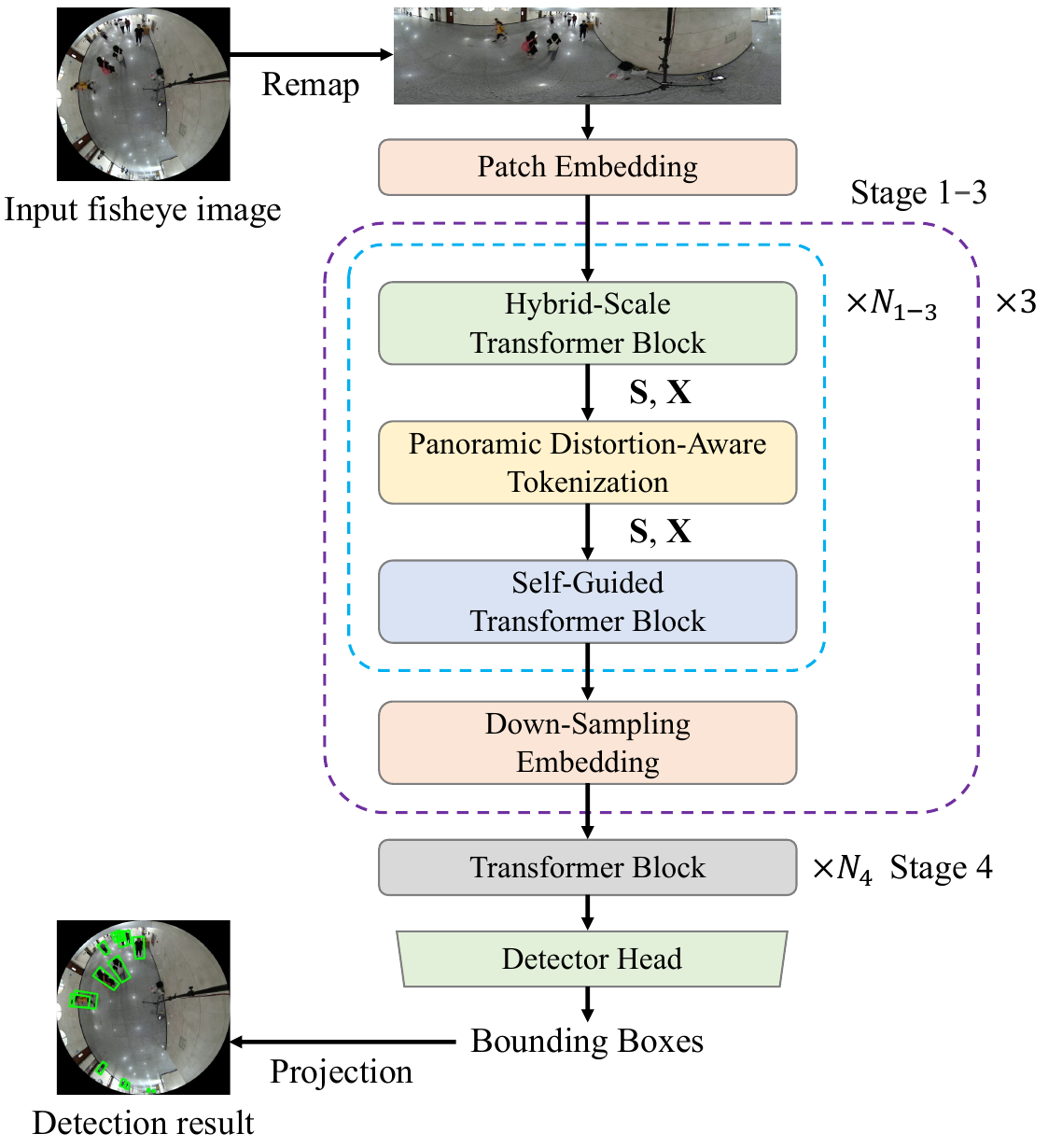}
\caption{Overall network architecture of our proposed method. The model consists of a detector head and a backbone with four stages. The $N$ in the subscript stage indicates the block count. The $\mathbf{S}$ and $\mathbf{X}$ are the significance map and feature map, respectively.}
\label{fig-architecture}
\end{figure}
Our network architecture consists of four main components: panoramic image conversion, a detector head based on the dynamic anchor box detection with transformers (DAB-DETR)~\cite{Liu_ICLR_2022}, a backbone based on SG-Former~\cite{Ren_ICCV_2023}, and our proposed PDAT.

\textbf{Panoramic image conversion.}
Our method converts an input fisheye image into an equirectangular panoramic image via remapping, as illustrated in~\fref{fig-architecture}. When the input image is rectangular, we center-crop the image. To address any uncalibrated cameras, we assume that the input images represent the stereographic projection, which is a standard fisheye camera model. In the case of radial distortion~\cite{Wakai_ECCV_2022}, this projection uses a trigonometric function: $\gamma = 2f_w \cdot \tan(\theta / 2)$, where $\gamma$ is the distortion, $f_w$ is the focal length, and $\theta$ represents the incident angle, as shown in \fref{fig-bbox-distribution}(a). We can then determine the focal length by fitting the image circle using $90^\circ$ incident angles. The image coordinates in the fisheye images are projected onto a unit sphere in world coordinates using backprojection~\cite{Wakai_ECCV_2022}, and then the world coordinates are projected onto an equirectangular image with an arbitrary image width~\cite{Sharma_AIVR_2019}. We use panoramic images with incident angles from $0^\circ$ to $90^\circ$ because the incident angle range from $90^\circ$ to $180^\circ$  represents the directions behind the cameras.

\textbf{Detector and backbone selection.}
Following the approach in~\cite{Yang_ICCV_2023}, we use DAB-DETR~\cite{Liu_ICLR_2022}, which has shown strong object detection performances. Window-based transformers are not suitable for use in the backbone of DAB-DETR because PDAT divides the feature maps nonuniformly. Therefore, we use an aggregation-based transformer called SG-Former~\cite{Ren_ICCV_2023} for the backbone. Note that SG-Former is the only aggregation-based method of this type available. The proposed network architecture is shown in \fref{fig-architecture}. The detector head block represents DAB-DETR without the backbone, and the remaining blocks represent SG-Former and PDAT. The patch embedding is extracted using the CNN, and the transformer block is a vanilla transformer block~\cite{Dosovitskiy_ICLR_2021}. The implementations of DAB-DETR and SG-Former follow the details provided in their respective papers. We explain the integration of PDAT into SG-Former in the following two steps: 1) SG-Former attention mechanism and 2) PDAT integration details.

\textbf{SG-Former attention mechanism.}
The core ideas of the SG-Former are hybrid-scale self-attention and self-guided attention. The hybrid-scale attention process extracts both global and local information from a feature map $\mathbf{X}\in \mathbb{R}^{H \times W \times C}$, where $H$, $W$, and $C$ are the height, the width, and the channels of the map, respectively. The hybrid-scale attention process also calculates the significance map $\mathbf{S}\in \mathbb{R}^{H \times W}$, based on the feature map. For example, the face is a high-significance area and the background is a low-significance area. The self-guided attention process aims to aggregate the feature vectors based on the significance map. For example, the feature vectors in a high-significance area of the face are aggregated into half the number of feature vectors; in contrast, the feature vectors in a low-significance area of the background are aggregated into 1/28th feature vectors. To obtain multi-scale features, down-sampling is applied between stages.

\textbf{PDAT integration details.}
We aim to extend the self-guided attention process using PDAT. Our modification of SG-Former inserts PDAT between the hybrid-scale transformer block and the self-guided block, as shown in \fref{fig-architecture}. The PDAT method leverages the significance map $\mathbf{S}$ based on the feature vectors grouped by the tiles. The integration process consists of the following steps. First, the significance maps are divided using our tiling method, as described in \sref{sec-panoramic-distortion-aware-tokenization}. From top to bottom, we split these maps horizontally into strips called regions, as shown in \fref{fig-our-tokenization}(b). Second, we extract 2D max pooling indices for each region $k$ using square kernels of $a_k \times a_k$ with horizontal strides $a_k$ in each tile. The $a_k$ corresponds to the height of region $k$ in the upper quarter of the image, with the exception that region 1 uses its half height, while a fixed height of $H/4$ is used for the remaining three-quarters. Finally, we leverage the significance values extracted by the max pooling from the second step. In this leveraging, the significance values are multiplied by the scale factor $\alpha$.

This approach seamlessly integrates into the self-guided attention procedure to preserve the significance values of smaller persons in the feature vector aggregation process. We introduce this scaling because larger persons dominate the attention mechanism by generating multiple high-significance areas in SG-Former without PDAT.

\subsection{Training and inference}
\label{sec-training-and-inference}
\textbf{Training.}
We generate equirectangular images from each training set. In this generation procedure, we avoid splitting the bounding boxes in the vertical image edges by changing the definition of the 0$^\circ$ azimuth angles. We then train our network using the panoramic images by following~\cite{Liu_ICLR_2022}.

\textbf{Inference.}
First, we convert an input fisheye image into a panoramic image by a remapping process, as illustrated in \fref{fig-architecture}. Second, the remapped panoramic image is then fed into our network. Finally, we project the detected bounding boxes onto the fisheye image, on which the shape of each bounding box is a rotated trapezoid. This projection represents the backprojection from a fisheye image to a panoramic image. The supplementary material describes details of the training and inference procedures.

\section{Experiments}
\label{sec-experiments}
\begin{table*}[t]
\caption{Person detection results using the average precision (AP) $\uparrow$ on the LOAF validation (val) and test sets}
\label{table-person-detection-result-on-loaf}
\centering
\scalebox{0.70}{
\begin{tabular}{cccc|ccc|cc||ccc|ccc|cc}  
\hline\noalign{\smallskip}
\multirow{2}{*}{Method} & \multicolumn{8}{c||}{$\texttt{val}$} & \multicolumn{8}{c}{$\texttt{test}$} \\
\cmidrule(lr){2-9}\cmidrule(lr){10-17}
~ & mAP & AP$_{50}$ & AP$_{75}$ & AP$_n$ & AP$_m$ & AP$_f$ & AP$_{seen}$ & AP$_{unseen}$  & mAP & AP$_{50}$ & AP$_{75}$ & AP$_n$ & AP$_m$ & AP$_f$ & AP$_{seen}$ & AP$_{unseen}$ \\
\noalign{\smallskip}
\hline\noalign{\smallskip}
Seidel~\etal{}~\cite{Seidel_VISAPP_2019} & 21.8 & 59.8 & \phantom{0}7.6 & 32.8 & 28.5 & \phantom{0}2.3 & 23.2 & 19.5 & 20.2 & 58.2 & \phantom{0}7.1 & 32.2 & 28.3 & \phantom{0}3.9 & 22.4 & 18.9 \\
Li~\etal{}~\cite{Li_AVSS_2019} & 28.5 & 63.3 & 20.1 & 46.8 & 24.2 & \phantom{0}1.3 & 33.8 & 29.3 & 27.2 & 65.2 & 21.3 & 47.2 & 24.8 & \phantom{0}1.3 & 31.8 & 27.6 \\
Tamura~\etal{}~\cite{Tamura_WACV_2019} & 34.8 & 72.1 & 27.7 & 51.7 & 38.8 & \phantom{0}8.7 & 38.8 & 32.9 & 34.2 & 72.8 & 28.7 & 53.7 & 37.3 & \phantom{0}6.5 & 39.5 & 33.2 \\
RAPiD~\cite{Duan_CVPRW_2020} & 40.3 & 77.9 & 34.8 & 55.3 & 41.9 & \phantom{0}9.2 & 44.7 & 37.6 & 39.2 & 77.9 & 35.4 & 54.8 & 40.1 & \phantom{0}7.9 & 44.2 & 37.3 \\
OARPD~\cite{Qiao_MTA_2024} & 45.5 & 78.8 & 46.3 & - & - & - & - & - & 45.8 & 79.4 & 45.8 & - & - & - & - & - \\
Yang~\etal{}~\cite{Yang_ICCV_2023} & 47.4 & 82.6 & 48.4 & 64.1 & 54.3 & 14.1 & 50.8 & 45.7 & 46.2 & 81.1 & 47.3 & \textbf{66.1} & 53.5 & 12.6 & 49.3 & 44.9 \\
\noalign{\smallskip}
\hline
\noalign{\smallskip}
Ours & \textbf{52.4} & \textbf{84.6} & \textbf{57.6} & \textbf{66.3} & \textbf{57.9} & \textbf{42.9} & \textbf{56.3} & \textbf{50.7} & \textbf{54.5} & \textbf{87.2} & \textbf{61.1} & 62.8 & \textbf{60.5} & \textbf{44.6} & \textbf{56.2} & \textbf{53.6}\\
\hline\noalign{\smallskip}
\end{tabular}
}
\end{table*}
\begin{table*}[t]
\caption{Person detection results on the CEPDOF and WEPDTOF datasets}
\label{table-person-detection-result-on-cepdof-and-wepdtof-datasets}
\centering
\scalebox{0.70}{
\begin{tabular}{ccccc|ccc||ccc|ccc}  
\hline\noalign{\smallskip}
\multicolumn{2}{c}{\multirow{2}{*}{Method}} & \multicolumn{6}{c||}{CEPDOF} & \multicolumn{6}{c}{WEPDTOF} \\
\cmidrule(lr){3-8}\cmidrule(lr){9-14}
~ & ~ & mAP & AP$_{50}$ & AP$_{75}$ & Precision & Recall & F1-score & mAP & AP$_{50}$ & AP$_{75}$ & Precision & Recall & F1-score \\
\noalign{\smallskip}
\hline\noalign{\smallskip}
Seidel~\etal{}~\cite{Seidel_VISAPP_2019} & \multicolumn{1}{c|}{VISAPP'19} & 20.9 & 50.6 & 10.2 & 80.6 & 39.5 & 53.0 & 16.1 & 39.4 & \phantom{0}9.0 & 70.9 & 38.6 & 50.0 \\
Li~\etal{}~\cite{Li_AVSS_2019} & \multicolumn{1}{c|}{AVSS'19} & 34.2 & 75.7 & 28.6 & 86.3 & 65.4 & 74.4 & 25.2 & 69.9 & 30.2 & 81.4 & 64.5 & 72.0 \\
Tamura~\etal{}~\cite{Tamura_WACV_2019}~$^1$ & \multicolumn{1}{c|}{WACV'19} & 28.8 & 61.6 & 23.0 & 73.1 & 65.8 & 69.3 & 12.8 & 27.2 & 10.9 & 53.0 & 34.3 & 41.6 \\
RAPiD~\cite{Duan_CVPRW_2020}~$^1$ & \multicolumn{1}{c|}{CVPRW'20} & 38.3 & 78.6 & 30.8 & \textbf{91.8} & 80.3 & 85.7 & 25.0 & 54.0 & 17.4 & \textbf{83.7} & 57.6 & 68.2 \\
OARPD~\cite{Qiao_MTA_2024}~$^1$ & \multicolumn{1}{c|}{MTA'24} & 11.4 & 28.9 & \phantom{0}6.6 & 80.5 & 31.3 & 45.1 & \phantom{0}4.1 & 10.0 & \phantom{0}2.2 & 28.8 & 19.0 & 22.9 \\
Yang~\etal{}~\cite{Yang_ICCV_2023}~$^1$ & \multicolumn{1}{c|}{ICCV'23} & 39.7 & 78.3 & 37.1 & 82.9 & 72.6 & 77.4 & 20.5 & 45.3 & 16.3 & 66.8 & 43.0 & 52.3 \\
\noalign{\smallskip}\hline\noalign{\smallskip}
\multicolumn{2}{c|}{Ours} & \textbf{42.8} & \textbf{85.6} & \textbf{38.7} & \textbf{91.8} & \textbf{81.8} & \textbf{86.5} & \textbf{37.5} & \textbf{74.9} & \textbf{34.8} & 81.7 & \textbf{69.0} & \textbf{74.8} \\
\hline\noalign{\smallskip}
\multicolumn{14}{l}{~$^1$ We trained networks for evaluation.} \\
\end{tabular}
}
\end{table*}
To demonstrate the validity and effectiveness of our approach, we conducted extensive experiments using three large-scale datasets~\cite{Duan_CVPRW_2020, Tezcan_WACV_2022, Yang_ICCV_2023}, which are composed of overhead fisheye images. We used 1024$\times$1024 input resolution by following the settings given in~\cite{Duan_CVPRW_2020, Li_AVSS_2019, Seidel_VISAPP_2019, Tamura_WACV_2019, Yang_ICCV_2023}. For evaluation, we converted the trapezoid bounding boxes into rectangles because no other methods listed in \tref{table-comparison-of-related-methods} produced trapezoid outputs. The experimental details are given in the supplementary material.

\subsection{Parameter settings}
\label{sec-parameter-settings}
For our network backbone, we used SG-Former-S~\cite{Ren_ICCV_2023}, which was pretrained on ImageNet~\cite{Russakovsky_IJCV_2015}. The SG-Former-S, which has the smallest backbone size among S, M, and B types, is comparable to Swin Transformer-T~\cite{Liu_ICCV_2021} used in Yang~\etal{}’s method~\cite{Yang_ICCV_2023}. We trained our network using a combination of a mini-batch size of 8 for 120 epochs and the AdamW optimizer~\cite{Loshchilov_ICLR_2019}. The initial learning rate was set to $2 \times 10^{-4}$ and was multiplied by 0.1 at the 40th epoch. We set the number of regions $K=5$ and the scale factor $\alpha=2$. Because the majority of bounding boxes were near the image circles of fisheye images, we adopted the 3072-pixel width of panoramic images, approximately corresponding to the circumference of fisheye images with 1024$\times$1024 pixels: $\pi \times 1024 = 3217 \approx 3072$ pixels. Unless otherwise specified, we employed panoramic images with 3072$\times$768 pixels. We provide implementation details and computational costs for different backbones in the supplementary.

\subsection{Person localization}
\label{sec-experiment-person-localization}
For person localization, we adopt the geometry-based method proposed by Yang~\etal{}~\cite{Yang_ICCV_2023}. This method estimates person locations by calculating the intersection point between the incident ray from the camera and the ground plane, where the incident ray is determined from the center of the lower edge of a detected person-bounding box.

\subsection{Experimental results}
\label{sec-experimental-resuls}
We used the person detection and localization results reported by Yang~\etal{}~\cite{Yang_ICCV_2023} in the methods of Seidel~\etal{}~\cite{Seidel_VISAPP_2019}, Li~\etal{}~\cite{Li_AVSS_2019}, Tamura~\etal{}~\cite{Tamura_WACV_2019}, RAPiD~\cite{Duan_CVPRW_2020}, and Yang~\etal{}~\cite{Yang_ICCV_2023}. We also used the OARPD~\cite{Qiao_MTA_2024} results reported in~\cite{Qiao_MTA_2024}. Unless otherwise specified, we refer to these results throughout. Because of the lack of previous performance reports, we implemented methods for comparison of Tamura~\etal{}~\cite{Tamura_WACV_2019} and OARPD~\cite{Qiao_MTA_2024}, based on the corresponding papers using PyTorch~\cite{Paszke_NIPS_2019}. In addition, the official codes from Yang~\etal{}~\cite{Yang_ICCV_2023} and RAPiD~\cite{Duan_CVPRW_2020} were used for evaluation. A qualitative evaluation is given in the supplementary material.

\subsubsection{Person detection on LOAF}
\label{sec-person-detection-on-loaf}
\begin{table}[t]
\caption{Ablation study on the LOAF test set}
\label{table-ablation-study}
\centering
\scalebox{0.70}{
\begin{tabular}{l|c@{\hspace{2mm}}c@{\hspace{2mm}}c@{\hspace{2mm}}|c@{\hspace{2mm}}c@{\hspace{2mm}}c@{\hspace{2mm}}|c@{\hspace{2mm}}c@{\hspace{2mm}}}
\hline\noalign{\smallskip}
~~~~~Method & mAP & AP$_{50}$ & AP$_{75}$ & AP$_{n}$ & AP$_{m}$ & AP$_{f}$ & AP$_{seen}$ & AP$_{unseen}$ \\
\noalign{\smallskip}\hline\noalign{\smallskip}
\phantom{$+$}~Baseline~$^1$ & 40.2 & 74.0 & 39.6 & 48.2 & 49.5 & 32.7 & 42.5 & 38.7 \\
$+$~Panorama~$^2$ & 49.8 & 84.7 & 52.7 & 59.3 & 55.3 & 39.7 & 51.0 & 48.9 \\
$+$~PDAT & \textbf{54.5} & \textbf{87.2} & \textbf{61.1} & \textbf{62.8} & \textbf{60.5} & \textbf{44.6} & \textbf{56.2} & \textbf{53.6} \\
\hline\noalign{\smallskip}
\multicolumn{9}{l}{~$^1$ ``Baseline'' denotes DAB-DETR using fisheye images.} \\
\multicolumn{9}{l}{~$^2$ ``Panorama'' denotes panoramic conversion.} \\
\end{tabular}
}
\end{table}
\begin{table}[t]
\caption{Panoramic-image width analysis on the LOAF test set}
\label{table-panoramic-image}
\centering
\scalebox{0.70}{
\begin{tabular}{c|c@{\hspace{2mm}}c@{\hspace{2mm}}c@{\hspace{2mm}}|c@{\hspace{2mm}}c@{\hspace{2mm}}c@{\hspace{2mm}}|c@{\hspace{2mm}}c@{\hspace{2mm}}}
\noalign{\smallskip}\hline\noalign{\smallskip}
~~Image width~~ & mAP & AP$_{50}$ & AP$_{75}$ & AP$_{n}$ & AP$_{m}$ & AP$_{f}$ & AP$_{seen}$ & AP$_{unseen}$ \\
\noalign{\smallskip}
\hline\noalign{\smallskip}
2048 pixels & 47.0 & 81.3 & 49.2 & 58.3 & 52.0 & 35.5 & 48.5 & 46.0 \\
2560 pixels & 51.2 & 85.1 & 55.6 & 61.9 & 56.4 & 39.3 & 53.8 & 49.7 \\
3072 pixels & \textbf{54.5} & \textbf{87.2} & \textbf{61.1} & \textbf{62.8} & \textbf{60.5} & \textbf{44.6} & \textbf{56.2} & \textbf{53.6} \\
\hline\noalign{\smallskip}
\end{tabular}
}
\end{table}
\textbf{Metrics and protocols.}
To demonstrate the validity and the effectiveness of the proposed method on LOAF, we used the mean average precision (mAP) following the COCO evaluation~\cite{Lin_ECCV_2014}. We used the subscripts $n$, $m$, and $f$ to denote the near (0--10 m), mid-range (10--20 m), and distant (over 20 m) person positions, respectively, e.g., AP$_n$. LOAF also provides seen and unseen scenes for test splits using the subscripts $seen$ and $unseen$, respectively. We used the official splits of the training, validation (val), and test sets.

\textbf{Results.}
Our method achieved the highest mAP among the methods listed in~\tref{table-person-detection-result-on-loaf} when used on both the validation and test sets. The mAP obtained using our method was substantially higher than that obtained using Yang~\etal{}'s method~\cite{Yang_ICCV_2023} by values of 5.0 and 8.3 on the LOAF validation and test sets, respectively. In addition, the AP$_f$ value obtained using our method on the test set was notably larger than that of Yang~\etal{}'s method~\cite{Yang_ICCV_2023} by 32.0, where 67\% of the bounding boxes were labeled as $f$: over 20 m. Our method also achieved an AP$_{seen}$ of 56.2 and an AP$_{unseen}$ of 53.6 on the test set, thus outperforming the other methods. The substantial margin between our method and Yang~\etal{}'s method~\cite{Yang_ICCV_2023} on the unseen test set of 8.7 demonstrates the scene robustness.

\subsubsection{Person detection on CEPDOF}
\label{sec-person-detection-on-cepdof}
\begin{table*}[t]
\caption{Person localization results using position errors (PE) $\downarrow$ in meters on the LOAF validation and test sets}
\label{table-person-localization}
\centering
\scalebox{0.72}{
\begin{tabular}{ccc|ccc|cc||c|ccc|cc}
\hline\noalign{\smallskip}
\multicolumn{2}{c}{\multirow{2}{*}{Method}} & \multicolumn{6}{c||}{$\texttt{val}$} & \multicolumn{6}{c}{$\texttt{test}$} \\
\cmidrule(lr){3-8}\cmidrule(lr){9-14}
\multicolumn{2}{c}{~} & ~~~mPE~~~ & PE$_n$ & PE$_m$ & PE$_f$ & PE$_{seen}$ & PE$_{unseen}$ & ~~~mPE~~~ & PE$_n$ & PE$_m$ & PE$_f$ & PE$_{seen}$ & PE$_{unseen}$ \\
\noalign{\smallskip}
\hline\noalign{\smallskip}
Seidel~\etal{}~\cite{Seidel_VISAPP_2019} & \multicolumn{1}{c|}{VISAPP'19} & 1.298 & 0.561 & 1.332 & 3.109 & 1.206 & 1.382 & 1.321 & 0.706 & 1.309 & 3.482 & 1.306 & 1.386 \\
Li~\etal{}~\cite{Li_AVSS_2019} & \multicolumn{1}{c|}{AVSS'19} & 0.898 & 0.502 & 0.871 & 2.650 & 0.832 & 0.962 & 0.913 & 0.543 & 0.884 & 2.780 & 0.904 & 0.998 \\
Tamura~\etal{}~\cite{Tamura_WACV_2019} & \multicolumn{1}{c|}{WACV'19} & 0.755 & 0.429 & 0.736 & 1.862 & 0.709 & 0.821 & 0.778 & 0.471 & 0.826 & 2.160 & 0.724 & 0.836 \\
RAPiD~\cite{Duan_CVPRW_2020} & \multicolumn{1}{c|}{CVPRW'20} & 0.674 & 0.426 & 0.623 & 1.403 & 0.625 & 0.757 & 0.682 & 0.461 & 0.664 &1.445 & 0.638 & 0.776 \\
OARPD~\cite{Qiao_MTA_2024} & \multicolumn{1}{c|}{MTA'24} & - & - & - & - & - & - & - & - & - & - & - & - \\
Yang~\etal{}~\cite{Yang_ICCV_2023} & \multicolumn{1}{c|}{ICCV'23} & 0.368 & 0.144 & 0.363 & 0.775 & 0.318 & 0.402 & 0.374 & 0.148 & 0.375 & 0.819 & 0.326 & 0.398 \\
\hline\noalign{\smallskip}
\multicolumn{2}{c|}{Ours} & \textbf{0.286} & \textbf{0.096} & \textbf{0.182} & \textbf{0.635} & \textbf{0.256} & \textbf{0.306} & \textbf{0.277} & \textbf{0.115} & \textbf{0.261} & \textbf{0.528} & \textbf{0.282} & \textbf{0.274} \\
\hline\noalign{\smallskip}
\end{tabular}
}
\end{table*}
\textbf{Metrics and protocols.}
CEPDOF~\cite{Duan_CVPRW_2020} provides bounding box annotations without person locations in the world coordinates. Therefore, we evaluated the detection performance using the AP. Similar to prior works~\cite{Duan_CVPRW_2020, Tezcan_WACV_2022, Yang_ICCV_2023}, we also used precision, recall, and the F1-score. Following the procedure of~\cite{Yang_ICCV_2023}, we trained the method of Tamura~\etal{}~\cite{Tamura_WACV_2019}, along with RAPiD~\cite{Duan_CVPRW_2020}, OARPD~\cite{Qiao_MTA_2024}, Yang~\etal{}~\cite{Yang_ICCV_2023}, and our method, using HABBOF~\cite{Li_AVSS_2019} and WEPDTOF~\cite{Tezcan_WACV_2022} after pretraining these methods on COCO~\cite{Lin_ECCV_2014}, although MW-R~\cite{Duan_CVPRW_2020} was not used for the training because it is not publicly available.

\textbf{Results.}
Our method achieved the highest mAP, AP$_{50}$, and AP$_{75}$ values on CEPDOF among the methods listed in~\tref{table-person-detection-result-on-cepdof-and-wepdtof-datasets}. The mAP, AP$_{50}$, and AP$_{75}$ values achieved using our method are higher than those obtained using Yang~\etal{}'s method~\cite{Yang_ICCV_2023} by values of 3.1, 7.3, and 1.6, respectively. Additionally, the F1-score of our method was greater than that of Yang~\etal{}'s method~\cite{Yang_ICCV_2023} by 9.1.

\subsubsection{Person detection on WEPDTOF}
\label{sec-person-detection-on-wepdtof}
\textbf{Metrics and protocols.}
We used the evaluation metrics of CEPDOF, as described in \sref{sec-person-detection-on-cepdof}. We trained the methods of Tamura~\etal{}~\cite{Tamura_WACV_2019}, RAPiD~\cite{Duan_CVPRW_2020}, OARPD~\cite{Qiao_MTA_2024}, Yang~\etal{}~\cite{Yang_ICCV_2023}, and our method, using HABBOF and CEPDOF after they were pretrained on COCO following~\cite{Yang_ICCV_2023}, although MW-R was again not used for the training because it is not publicly available.

\textbf{Results.}
Similar to the CEPDOF results, our method achieved the highest mAP, AP$_{50}$, and AP$_{75}$ results on WEPDTOF among the methods listed in~\tref{table-person-detection-result-on-cepdof-and-wepdtof-datasets}. The mAP values achieved on CEPDOF using both the conventional methods and our method were larger than those obtained on WEPDTOF because WEPDTOF has more scenes; \ie, WEPDTOF consists of 16 videos, and CEPDOF consists of 8 videos. Comparison between the CEPDOF and WEPDTOF results shows that the methods of Tamura~\etal{}~\cite{Tamura_WACV_2019}, RAPiD~\cite{Duan_CVPRW_2020}, OARPD~\cite{Qiao_MTA_2024}, and Yang~\etal{}~\cite{Yang_ICCV_2023} apparently require diverse images because the network is trained on fisheye images that include rotated persons.

\subsubsection{Diagnostic study}
\label{sec-diagnostic-study}
We conducted a diagnostic study on the LOAF test set. \tref{table-ablation-study} shows the ablation study results. The baseline represents DAB-DETR~\cite{Liu_ICLR_2022} with the SG-Former-S~\cite{Ren_ICCV_2023} for the backbone when using fisheye input images. The second row demonstrates the effectiveness of the panoramic conversion. The panoramic images were suitable for DAB-DETR because they removed the person rotations. After the panoramic conversion was adopted, the performance increased by an mAP of 9.6. We also validated the effectiveness of the PDAT process. The third row indicates that our method using PDAT provided substantially improved person detection results when compared with our method without PDAT by an mAP of 4.7. Therefore, the PDAT alleviated the problems caused by the variations in person size dramatically. Similar to the mAP, the other metrics were improved notably.

In addition, \tref{table-panoramic-image} shows the dependence of the panoramic image width. The 3072-pixel panoramic image width is suitable for the 1024$\times$1024-pixel fisheye images because the circumference of the fisheye image is $\pi \times 1024 = 3217 \approx 3072$. Therefore, the 3072-pixel image width without shrinkage led to the best performance among the widths of 2048, 2560, and 3072 pixels.

\subsubsection{Person localization}
\label{sec-person-localization}
\textbf{Metrics and protocols.}
To validate the person localization accuracy, we compared our method with the conventional methods. We evaluated the individual person positions from the bounding boxes using Yang~\etal{}'s approach~\cite{Yang_ICCV_2023}, as described in \sref{sec-experiment-person-localization}. We used the position error (PE) in meters as the horizontal distance between the ground-truth person position and the person position estimated from the bounding boxes. We also used the same subscripts used in \sref{sec-person-detection-on-loaf}, \eg{}, PE$_n$. These estimated bounding boxes were the same as the person detection results on LOAF from \sref{sec-person-detection-on-loaf}.

\textbf{Results.}
Our method achieved the lowest mean PE (mPE), as indicated in~\tref{table-person-localization}. The mPE for our method is smaller than that of Yang~\etal{}'s method~\cite{Yang_ICCV_2023} on the LOAF test set by 0.097 m. Similar to the person detection performance, our method realized a noteworthy 0.124-m gain in terms of the PE$_{unseen}$ when compared with Yang~\etal{}'s method~\cite{Yang_ICCV_2023}. Our method outperformed the other methods on PE$_n$, PE$_m$, and PE$_f$ because of the accuracy of bounding box detection.

\section{Conclusion}
\label{sec-conclusion}
\textbf{Limitations.}
We assume that the fisheye images are stereographic projections. However, this assumption may cause a mismatch with the actual distortion. To address this problem, we will extend our method using deep single image camera calibration techniques~\cite{Dal_ECCV_2024, Wakai_ECCV_2022} that estimate the lens distortion. Another promising direction for future work is to use videos for the input. In this paper, we have focused on person detection and localization from an image. 

We have proposed a transformer-based detection and localization method for use with overhead fisheye images. Panoramic remapping and PDAT address person rotation and small-person problems in overhead fisheye images. Our nonuniform tokenization using self-similar figures leverages the significance values to alleviate the imbalance in the significance areas of the feature maps. Experiments demonstrated that our method outperforms the prior methods.

\clearpage

\setcounter{section}{0}

\twocolumn[{
    \vspace{10mm}
    \centering
        \Large
        \textbf{\thetitle}\\
        \textbf{Supplementary}
        \vspace{7mm}

    \centerline{
    \fontsize{12pt}{0cm}\selectfont
        Nobuhiko Wakai$^1$ \quad\quad Satoshi Sato$^1$ \quad\quad Yasunori Ishii$^1$ \quad\quad Takayoshi Yamashita$^2$
    }

    \centerline{
    \fontsize{12pt}{0cm}\selectfont
        $^1$ Panasonic Holdings Corporation\quad\quad $^2$ Chubu University
    }

    {\tt\small \{wakai.nobuhiko,sato.satoshi,ishii.yasunori\}@jp.panasonic.com} \quad {\tt\small takayoshi@isc.chubu.ac.jp}
    \vspace{11mm}
}]

\textbf{Structure of this paper.}
In this supplementary material, we present some details omitted from the main paper. For our proposed method in \sref{sec-proposed-method} (main paper), we demonstrate distribution of bounding boxes in~\sref{sec-distribution-of-bounding-boxes}, details of the panoramic distortion-aware tokenization (PDAT) in~\sref{sec-details-of-panoramic-distortion-aware-tokenization}, and implementation details in~\sref{sec-implementation-details}. For the experiments in \sref{sec-experiments} (main paper), we describe computational costs in~\sref{sec-computational-costs}, the experimental details in~\sref{sec-experimental-details}, and qualitative evaluation in~\sref{sec-qualitative-evaluation}. Furthermore, we provide extended related work in~\sref{sec-extended-related-work}, the novelty in~\sref{sec-novelty}, and the broader impact in~\sref{sec-broader-impact}.

\section{Distribution of bounding boxes}
\label{sec-distribution-of-bounding-boxes}
We present the detailed description of the bounding box distribution in an overhead fisheye dataset, as shown in \fref{fig-bbox-distribution} (main paper).

\subsection{Relationship between bounding box height and incident angle}
\label{sec-relationship-between-bounding-box-height-and-incident-angle}
\begin{figure*}[t]
\centering
\includegraphics[width=0.70\hsize]{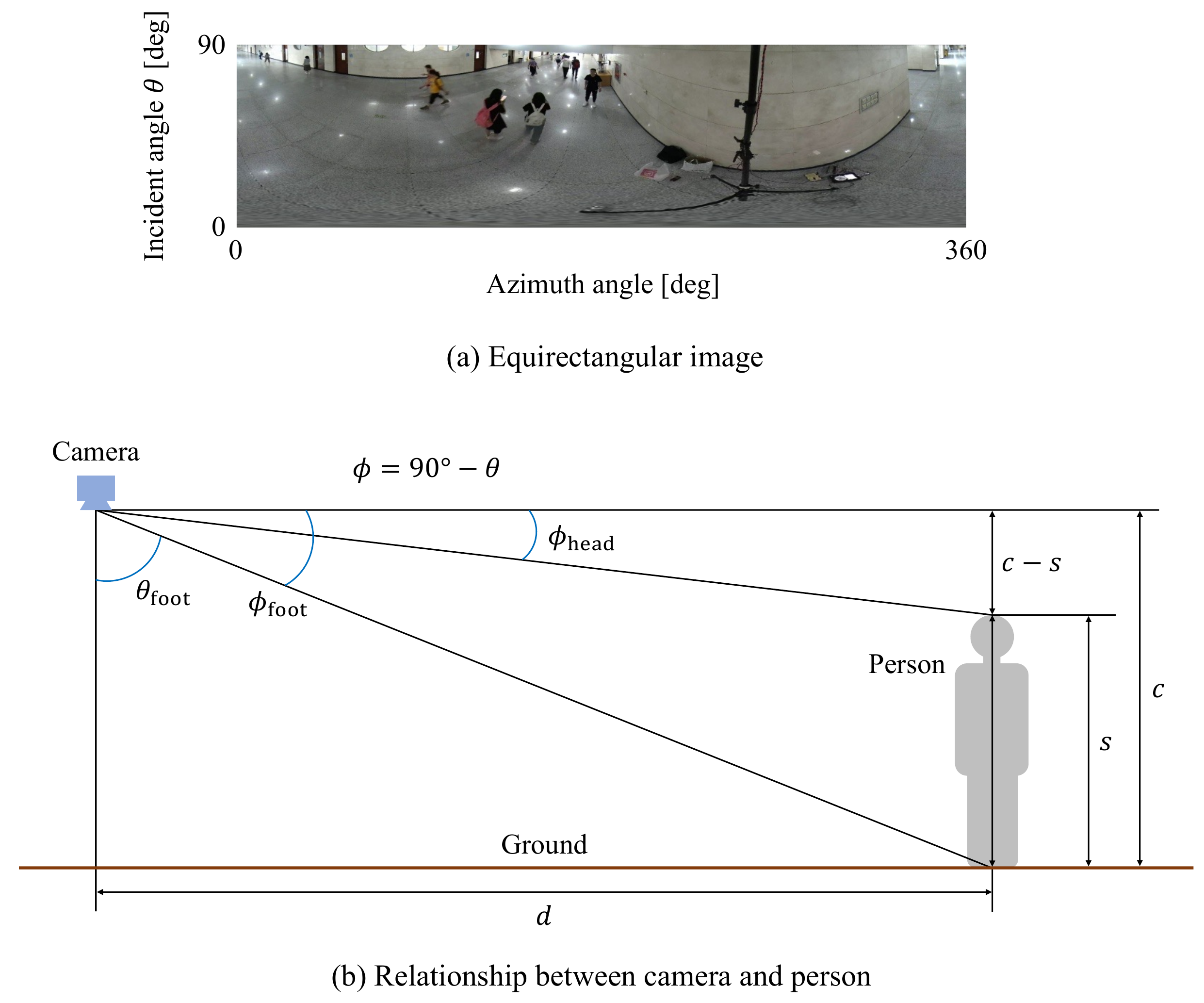}
\caption{Definition of angles and lengths. (a) An equirectangular image. The vertical axis shows the incident angle, ranging from 0$^\circ$ (nadir) to 90$^\circ$ (horizon), and the horizontal axis shows the azimuth angle. An overhead fisheye image in LOAF~\cite {Yang_ICCV_2023} was converted to the equirectangular image. (b) Geometric relationship between an overhead fisheye camera and a person. The parameters are defined as follows: $c$ denotes the camera height, $s$ represents the person's height, and $d$ indicates the horizontal distance from the camera to the person. The angle $\phi$ is measured from the horizontal axis, and $\theta$ represents the incident angle, where $\phi=90^\circ-\theta$, \eg{}, $\phi_{\rm{foot}}=90^\circ-\theta_{\rm{foot}}$.}
\label{fig-derivation-of-box}
\end{figure*}
\begin{figure*}[t]
\centering
\includegraphics[width=0.90\hsize]{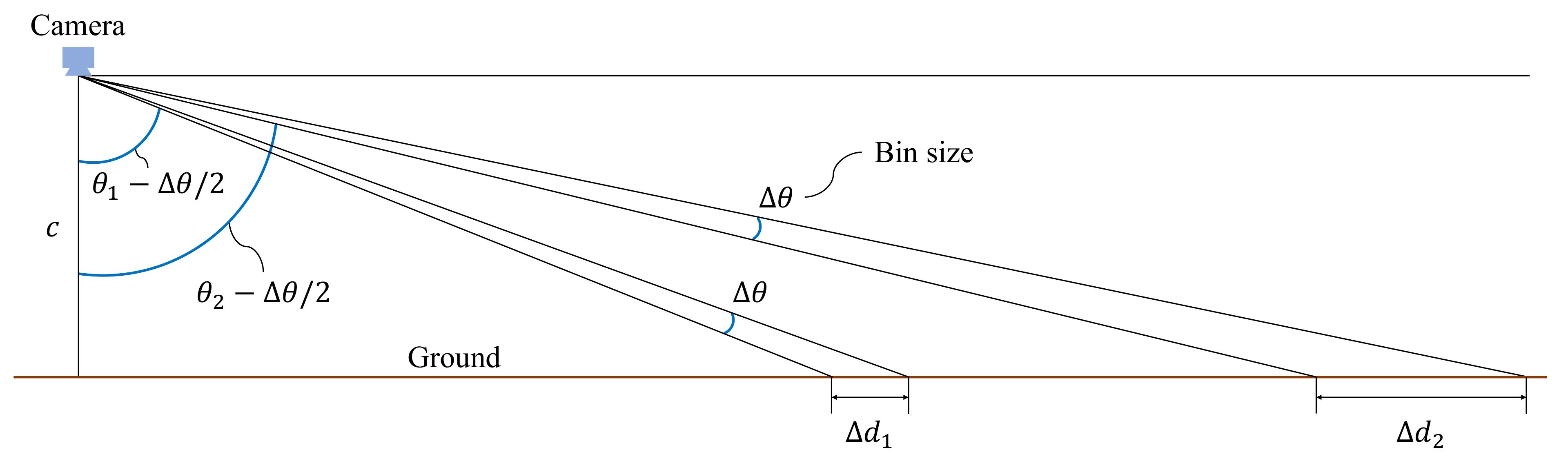}
\caption{Geometric relationship between incident angles $\theta$ and horizontal distances $\Delta d$ for equal angular intervals $\Delta\theta$: the bin size of our distribution analysis.}
\label{fig-derivation-of-small-person}
\end{figure*}
We demonstrate that the height of the bounding boxes changes approximately linearly with the incident angle when the incident angle is large enough ($>70^\circ$). \fref{fig-derivation-of-box} shows the definition of angles and lengths based on the camera and a person position in the world coordinates. To simplify the conditions, we assume that a person stands at a distance $d$ from the camera. The heights of the camera and the person are represented by $c$ and $s$, respectively. We describe the relationship between the bounding box height and the incident angle. First, we obtain the relationships among the values of $c$, $s$, and $d$ as follows:
\begin{align}\label{eq-step-1}
\left\{
\begin{array}{cc}
c -s & = d \tan \phi_{{\rm{head}}} \\
c    & = d \tan \phi_{{\rm{foot}}} \\
\end{array}
\right.,\quad c, s, d > 0,\quad c > s,
\end{align}
where $\phi$ represents the angles formed between the points of the person and the horizontal plane at the camera; the subscripts $\rm{head}$ and $\rm{foot}$ refer to the points of the head and the foot of the person, as shown in \fref{fig-derivation-of-box}. Note that $\phi$ is equal to $90^\circ-\theta$, where $\theta$ is the incident angle. Second, by removing $d$ from \eref{eq-step-1}, we obtain the relationship between $\phi_{{\rm{head}}}$ and $\phi_{{\rm{foot}}}$ as shown:
\begin{equation}
\label{eq-step-2}
\tan \phi_{{\rm{foot}}} = \frac{c}{c -s } \tan \phi_{{\rm{head}}},~~ \frac{c}{c - s} > 1.
\end{equation}
Third, under practical conditions, the distance $d$ is sufficiently large; thus, $\phi_{{\rm{foot}}}$ and $\phi_{{\rm{head}}}$ are sufficiently small. Therefore, we assume that $\tan \phi_{{\rm{foot}}}$ is equal to $\phi_{{\rm{foot}}}$ and that $\tan \phi_{{\rm{head}}}$ is equal to $\phi_{{\rm{head}}}$ based on the first-order Taylor series expansions given by
\begin{equation}
\label{eq-step-3}
\phi_{{\rm{foot}}} = \frac{c}{c -s }~\phi_{{\rm{head}}}.
\end{equation}
Fourth, the bounding box height $B_H$ in the equirectangular panoramic images is given by $\phi_{{\rm{foot}}} - \phi_{{\rm{head}}}$ because the vertical coordinates of the equirectangular images are given by values of $\phi$ from $0^\circ$ to $90^\circ$ from the top of the image to the bottom, as follows:
\begin{equation}
\label{eq-step-4}
B_H = \lambda(\phi_{{\rm{foot}}} - \phi_{{\rm{head}}}),
\end{equation}
where $\lambda$ is a linear function that converts the angle unit into the pixel unit on the equirectangular image, depending on the focal length and image sensor pitches. Finally, using \eref{eq-step-3} and \eref{eq-step-4}, we obtain the relationship between the bounding box height $B_H$ and $\phi$ as follows:
\begin{equation}
\label{eq-step-5}
B_H = \frac{s}{c - s}~\lambda(\phi_{{\rm{head}}}) \propto \phi_{{\rm{head}}}.
\end{equation}
Note that the angle of the center of the person is approximately equal to $\phi_{{\rm{head}}}$ because the distance $d$ is sufficiently large. Therefore, the bounding box height increases linearly from the top of the equirectangular panoramic images to the bottom of these images when $\phi$ is small enough; \ie, persons are located at a distance from the camera. Therefore, we obtained the distribution of the bounding boxes, as shown in \fref{fig-bbox-distribution}(b) (main paper).

\subsection{Relationship between ground area and incident angle}
\label{sec-relationship-between-ground-area-and-incident-angle}
\begin{figure*}[t]
\centering
\includegraphics[width=0.95\hsize]{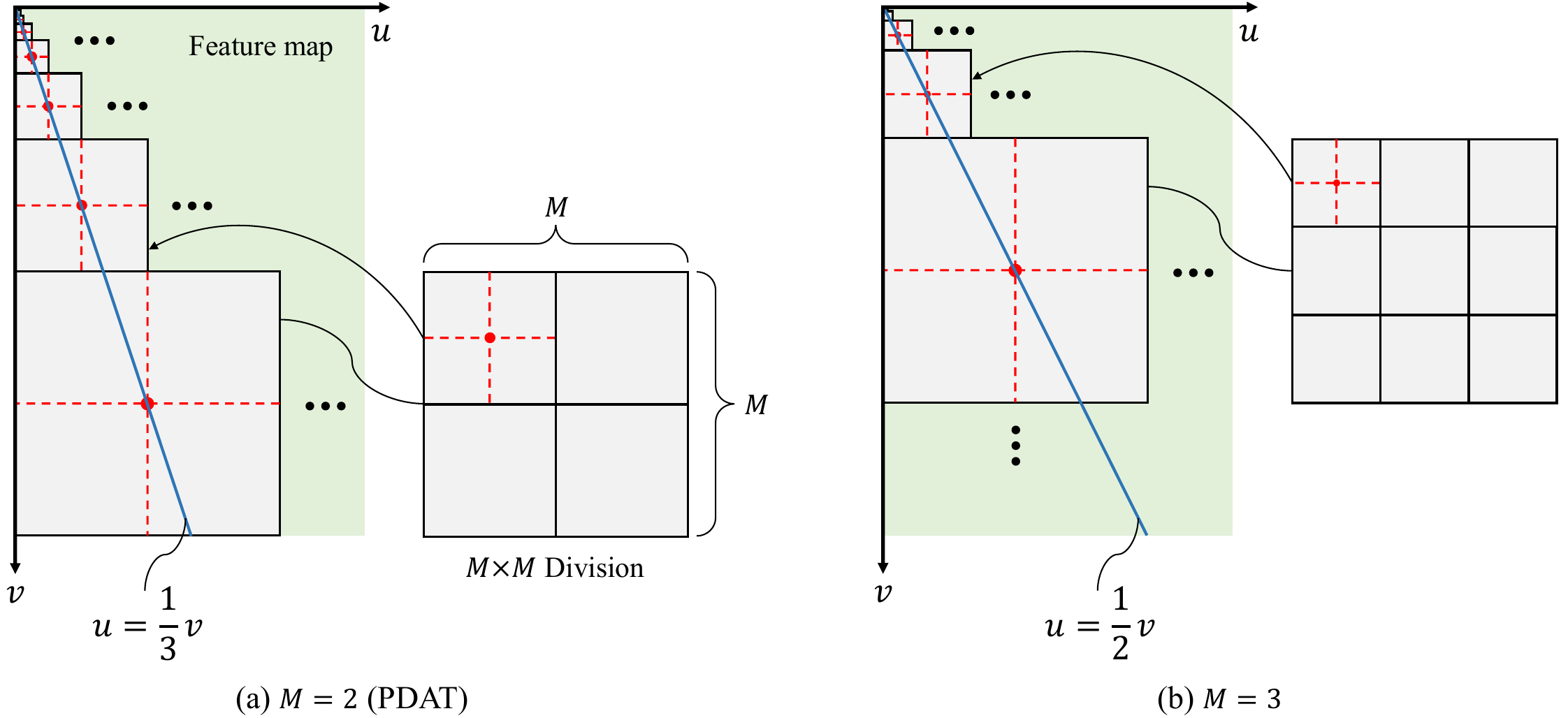}
\caption{Comparison of tokenization between panoramic distortion-aware tokenization (PDAT) and the division using trichotomy. (a) Division using PDAT ($M=2$). (b) Division using trichotomy ($M=3$). The green area indicates a feature maps $\mathbf{X}\in \mathbb{R}^{H \times W \times C}$, where $H$, $W$, and $C$ are the height, the width, and the channels of the map, respectively. The blue lines pass through the tile centers, following the relationship $u=v/3$ in (a), while $u=v/2$ in (b). The $u$ and $v$ axes are along $W$ and $H$ of the feature map, respectively.}
\label{fig-details-pdat}
\end{figure*}
As shown in \fref{fig-bbox-distribution}(b) (main paper), we observed that 75\% of all person-bounding boxes are concentrated in the range of incident angles from 80$^\circ$ to 90$^\circ$, indicating a strong bias in person distribution. As discussed in \sref{sec-geometric-analysis-of-overhead-fisheye-images} (main paper), this bias has two main causes: 1) peripheral image portions at large incident angles subtend larger ground areas, and 2) motion parallax~\cite{Gibson_JEP_1959, Hartley_CUP_2004}. Here, we provide a further description of the first cause.

\fref{fig-derivation-of-small-person} shows the horizontal distance $\Delta d$ formed by two incident angles of $\theta + \Delta\theta/2$ and $\theta - \Delta\theta/2$ as
\begin{equation}
\label{eq-delta_d}
\Delta d = c\,(\tan(\theta + \Delta\theta/2) - \tan(\theta - \Delta\theta/2)),
\end{equation}
where $c$ is the camera height and $\Delta\theta$ is the bin size of our distribution analysis: 1$^\circ$. Under $\theta_2 > \theta_1$, $\Delta d_2$ is longer than $\Delta d_1$ because of the prominent increase along $\theta$ of the tangent function. A larger horizontal distance $d$ corresponds to a larger ground area, which tends to include more persons. Note that incident angles primarily dominate this tendency at large incident angles because the effect of the oblique viewing angles of persons is negligible; \ie{}, the appearance changes of persons caused by the oblique viewing angles are small for distant persons. Additionally, as shown in \fref{fig-bbox-distribution}(b) (main paper), the number of bounding boxes within the $\theta$-range of $87^\circ-90^\circ$ decreases due to the lack of annotations; \ie{}, the person size is too small for annotators. Therefore, we observed the concentration of bounding boxes at large incident angles.

\section{Details of panoramic distortion-aware tokenization}
\label{sec-details-of-panoramic-distortion-aware-tokenization}
We describe the panoramic distortion-aware tokenization (PDAT) method in detail. In the PDAT approach, a panoramic image is divided into square tiles based on self-similar figures, as described in \sref{sec-panoramic-distortion-aware-tokenization} (main paper). The unit figure for PDAT consists of a square and two rectangles, and each rectangle can be divided recursively, as illustrated in \fref{fig-our-tokenization}(b) (main paper); \ie, the sequences involve the tile size decreasing continuously by half.

We examine another case in which the tile sizes are reduced continuously by dividing them into thirds, \ie, 3$\times$3 divisions. \fref{fig-details-pdat}(a) shows the division of PDAT using $M = 2$; \ie{}, the tile sizes are reduced continuously by dividing them into $M$$\times$$M$, where $M$ is the division number of the side of a tile. We found that a panoramic image can also be tiled using trichotomy ($M = 3$), which yields division results without gaps, as shown in \fref{fig-details-pdat}(b).

\fref{fig-details-pdat} shows that the coefficients of the linear function are $1/3$ and $1/2$ in (a) and (b), respectively. This linear function, passing through the tile centers, indicates that the tile size linearly changes along the vertical axis. In the general case of $M$, we describe the coefficient $w_M$ of a linear function of $u = w_M~v$, where $u$ and $v$ are the horizontal and vertical axes, respectively. We focused on two adjacent tiles along the $v$ axis. When the side of one tile is $b$, the side of the other is $Mb$. The coefficient $w_M$ can be calculated using the two centers of these tiles as shown:
\begin{equation}
\label{eq-pdat}
w_M = \left(\frac{Mb}{2}-\frac{b}{2}\right) / \left(\frac{b(M+2)}{2}-\frac{b}{2}\right) = \frac{M-1}{M+1}.
\end{equation}
In terms of transformer tokenization in person detection, the small difference in tile size between adjacent regions is desired because a large tile size difference may prevent our network from detecting bounding boxes that are located in two or more different regions, \eg, in regions 3 and 4. To minimize the difference, we used $M=2$ that minimizes $w_M$ $(M \ge 2)$ in \eref{eq-pdat}. Therefore, the PDAT method using $M=2$ represents the optimal division in terms of maintaining consistent person-to-tile size ratios, particularly in portions with large incident angles.

\section{Implementation details}
\label{sec-implementation-details}
We discuss the implementation details of the proposed method for reproducibility. Our method was implemented using PyTorch~\cite{Paszke_NIPS_2019}.

\subsection{Panoramic image conversion}
In terms of camera geometry, we describe the details of the panoramic conversion in \sref{sec-network-architecture} (main paper). Our method converts an input fisheye image into an equirectangular image by remapping. Note that we apply center-cropping to obtain a square image when the input image is rectangular. If the top and bottom of the circumferential fisheye image are truncated in the input image, we zero-pad the truncated areas to form a square image. To address uncalibrated cameras, we assume that the input images follow the stereographic projection model. This is one of the standard fisheye camera models, along with orthogonal projection, equisolid angle projection, and equidistance projection. Among these models, the stereographic projection provides the largest image areas in the peripheral portions, making it suitable for monitoring distant people in visual surveillance.

Camera models express the required mapping from world coordinates $\vph$ to image coordinates $\vuh$ in homogeneous coordinates. In the case of radial distortion~\cite{Wakai_ECCV_2022}, this mapping is represented as
\begin{align}\label{eq-nonlinear-model}
\vuh = 
\left[
\begin{array}{ccc}
\gamma / d_u & 0 & c_u \\
0 & \gamma / d_v & c_v \\
0 & 0 & 1
\end{array}
\right] \extrinsics \vph,
\end{align}
where $\gamma$ is the distortion, $\left(d_u, d_v\right)$ is the image sensor pitch, $\left(c_u, c_v\right)$ is a principal point, $\rot$ is a rotation matrix, and $\trans$ is a translation vector. The subscripts $u$ and $v$ denote the horizontal and vertical directions, respectively.

In \eref{eq-nonlinear-model}, the stereographic projection model represents $\gamma = 2f_w \cdot \tan(\theta / 2)$, where $f_w$ is the focal length, and $\theta$ is the incident angle, as shown in \fref{fig-derivation-of-box}(a). We can then determine the focal length by fitting the image circle using $90^\circ$ incident angles; \ie{}, the diameter of the image circle corresponds to the fisheye image height. Assuming $d_u = d_v$, which means the image sensor pixels are square, these sensor pitch parameters do not have a degree of freedom because $f_w$, $d_u$, and $d_v$ represent a single degree, which is the scale factor of the camera model. For the principal point $(c_u, c_v)$, we use the image center. The extrinsics $\extrinsics$ can be set arbitrarily because these parameters do not affect the conversion from fisheye to equirectangular images.

On the basis of the projection above, the image coordinates $\vuh$ in the fisheye images are backprojected onto a unit sphere in world coordinates $\vph$~\cite{Wakai_ECCV_2022}, and then the world coordinates are projected onto an equirectangular image with an arbitrary image width~\cite{Sharma_AIVR_2019}. We use panoramic images with incident angles ranging from $0^\circ$ to $90^\circ$, as shown in \fref{fig-derivation-of-box}(a), because the incident angle range from $90^\circ$ to $180^\circ$ represents the directions behind the cameras.

\subsection{SG-Former modification}
We described the details of PDAT integration for DAB-DETR and SG-Former in \sref{sec-network-architecture} (main paper). In addition to the description, we modified the position embedding of SG-Former to be interpolated for panoramic images following~\cite{Dosovitskiy_ICLR_2021} because the pretrain model was trained on ImageNet~\cite{Russakovsky_IJCV_2015}, whose resolution is 224$\times$224 pixels. Furthermore, we adjusted the window size to 8 instead of 7 because panoramic images with a 3072-pixel width cannot be evenly divided by 7. The hyperparameters of DAB-DETR and SG-Former were determined on the basis of \cite{Yang_ICCV_2023} and \cite{Ren_ICCV_2023}, respectively.

\subsection{Training and Inference}
We describe the details of our training and inference phases in \sref{sec-training-and-inference} (main paper).
\begin{table*}[t]
\caption{Computational costs of SG-Former-S and the detector head}
\label{table-Computational-costs-of-SG-Former-S-and-the-detector-head}
\centering
\scalebox{0.88}{
\begin{tabular}{cccccccc}
\hline\noalign{\smallskip}
\multirow{2}{*}{Module} & \multirow{2}{*}{~~~\#Params~$^{1}$~~} & \multicolumn{3}{c}{GFLOPs~$^{1}$} & \multirow{2}{*}{\#Blocks ($N$)} & \multirow{2}{*}{Size~$^3$} & \multirow{2}{*}{\#Channels ($C$)} \\
\cmidrule(lr){3-5}
~ & ~ & W3072~$^2$ & ~W2560~ & ~W2048~ & ~ & ~ & ~ \\
\noalign{\smallskip}
\hline
\noalign{\smallskip}
\multicolumn{1}{l|}{Backbone: SG-Former-S} & \multicolumn{1}{c|}{21.83M} & 214.3 & 148.8 & \multicolumn{1}{c|}{\phantom{0}95.3} & - & - & - \\
\multicolumn{1}{l|}{~~~Patch embedding} & \multicolumn{1}{c|}{\phantom{0}0.08M} & \phantom{0}28.4 & \phantom{0}19.7 & \multicolumn{1}{c|}{\phantom{0}12.6} & - & $W/4 \times H/4$ & \phantom{0}64 \\
\multicolumn{1}{l|}{~~~Stage 1} & \multicolumn{1}{c|}{\phantom{0}0.38M} & \phantom{0}15.6 & \phantom{0}10.8 & \multicolumn{1}{c|}{\phantom{00}6.9} & 2 & $W/4 \times H/4$ & \phantom{0}64 \\
\multicolumn{1}{l|}{~~~Down-sampling embedding 1} & \multicolumn{1}{c|}{\phantom{0}0.03M} & \phantom{00}1.9 & \phantom{00}1.3 & \multicolumn{1}{c|}{\phantom{00}0.8} & - & $W/8 \times H/8$ & 128 \\
\multicolumn{1}{l|}{~~~Stage 2} & \multicolumn{1}{c|}{\phantom{0}1.41M} & \phantom{0}30.3 & \phantom{0}21.0 & \multicolumn{1}{c|}{\phantom{0}13.5} & 4 & $W/8 \times H/8$ & 128 \\
\multicolumn{1}{l|}{~~~Down-sampling embedding 2} & \multicolumn{1}{c|}{\phantom{0}0.10M} & \phantom{00}1.9 & \phantom{00}1.3 & \multicolumn{1}{c|}{\phantom{00}0.8} & - & $W/16 \times H/16$ & 256 \\
\multicolumn{1}{l|}{~~~Stage 3} & \multicolumn{1}{c|}{\textbf{15.99M}} & \textbf{126.5} & \phantom{0}\textbf{87.9} & \multicolumn{1}{c|}{\phantom{0}\textbf{56.2}} & \textbf{16} & $W/16 \times H/16$ & 256 \\
\multicolumn{1}{l|}{~~~Down-sampling embedding 3} & \multicolumn{1}{c|}{\phantom{0}0.40M} & \phantom{00}1.8 & \phantom{00}1.3 & \multicolumn{1}{c|}{\phantom{00}0.8} & - & $W/32 \times H/32$ & 512 \\
\multicolumn{1}{l|}{~~~Stage 4} & \multicolumn{1}{c|}{\phantom{0}3.44M} & \phantom{00}7.9 & \phantom{00}5.5 & \multicolumn{1}{c|}{\phantom{00}3.5} & 1 & $W/32 \times H/32$ & 512 \\
\noalign{\smallskip}
\hline\hline
\noalign{\smallskip}
\multicolumn{1}{l|}{Detector head: DAB-DETR} & \multicolumn{1}{c|}{20.17M} & 409.7 & 285.2 & \multicolumn{1}{c|}{183.3} & - & - & - \\
\hline\noalign{\smallskip}
\multicolumn{8}{l}{\scalebox{1.00}{~$^1$ Computational costs were calculated using~\cite{MMCV_MISC_2018}.}} \\
\multicolumn{8}{l}{\scalebox{1.00}{~$^2$ W3072, W2560, and W2048 denote panoramic image widths of 3072, 2560, and 2048 pixels, respectively.}} \\
\multicolumn{8}{l}{\scalebox{1.00}{~$^3$ ``Size'' denotes the spatial dimensions of feature maps relative to the input image of size $W \times H$.}} \\
\end{tabular}
}
\end{table*}

\subsubsection{Details of the training}
\label{sec-details-of-the-training}
We generated equirectangular panoramic images from each training set to train our networks. This generation is the reverse process of converting panoramic images into fisheye images in deep single image camera calibration~\cite{Wakai_ECCV_2022, Wakai_CVPR_2024}. We avoided splitting the bounding boxes in the vertical image edges by changing the definition of the 0$^\circ$ azimuth angles. Following the DAB-DETR training procedure, we trained our network using SG-Former, which had been pretrained on ImageNet~\cite{Russakovsky_IJCV_2015}. We used the AdamW optimizer~\cite{Loshchilov_ICLR_2019} whose weight decay is $1 \times 10^{-4}$ for 120 epochs, with a mini-batch size of 8 on the gradient accumulation. The initial learning rate of DAB-DETR was set to $2 \times 10^{-4}$ and this rate was multiplied by 0.1 at the 40th epoch. The constant learning rate of SG-Former was set to $2 \times 10^{-5}$. Note that we pretrained our network using the COCO dataset~\cite{Lin_ECCV_2014} for CEPDOF~\cite{Duan_CVPRW_2020} and WEPDTOF~\cite{Tezcan_WACV_2022} evaluation only in a manner similar to comparison methods. This pretraining was conducted for 50 epochs using only person labels, which are not crowded (``iscrowd'' is false).

On the basis of the procedures in~\cite{Yang_ICCV_2023} and~\cite{Liu_ICLR_2022}, we conducted data augmentations for overhead fisheye image datasets and the COCO dataset, respectively. Instead of rotation augmentation for overhead fisheye images~\cite{Yang_ICCV_2023}, we employed the random horizontal circular shift on panoramic images that corresponds to image rotations on fisheye images. In random resizing and cropping augmentation, we maintained the top of panoramic images; \ie, the incident angle of 90$^\circ$ corresponded to the top edge of images for PDAT. These modifications of data augmentations were required by the difference in input images between panoramic and fisheye images.

Our training platform was equipped with an Intel Core i9-13900K CPU and an NVIDIA GeForce RTX 4090 GPU with 24 GB of GPU memory. The training time was approximately 9 hours per epoch on the LOAF dataset. To reduce GPU memory usage, we used the techniques of automatic multi-precision of FP16 and gradient checkpoints, which were provided by PyTorch~\cite{Paszke_NIPS_2019}.

\subsubsection{Details of the inference}
First, an input fisheye image was converted into an equirectangular panoramic image within the incident angles from $0^\circ$ to $90^\circ$ by a remapping method. In this remapping, we selected the definition of 0$^\circ$ azimuth angles at random. Second, the converted panoramic image was fed into our network. Following~\cite{Yang_ICCV_2023}, we did not use test-time augmentation. Finally, we projected the rectangular bounding boxes from the panoramic image onto the fisheye image, resulting in rotated trapezoids. For evaluation, we converted the trapezoid bounding boxes into rectangles. The width of the rectangle was calculated using the mean of the parallel sides of the trapezoid. For inference, we used the same machine that was used for the training, as described in \sref{sec-details-of-the-training}. The inference computational cost is explained in \sref{sec-computational-costs}.
\begin{table*}[t]
\begin{minipage}[t]{\textwidth}
\centering
\caption{Computational costs of SG-Former-T using panoramic image width of 2048 pixels}
\label{table-Computational-costs-of-SG-Former-T-using-panoramic-image-width-of-2048-pixels}
\centering
\scalebox{0.88}{
\begin{tabular}{cccccccc}
\hline\noalign{\smallskip}
\multirow{2}{*}{Module} & \multicolumn{2}{c}{With global~$^1$} & \multicolumn{2}{c}{Without global~$^1$} & \multirow{2}{*}{\#Blocks ($N$)} & \multirow{2}{*}{Size~$^3$} & \multirow{2}{*}{\#Channels ($C$)} \\
\cmidrule(lr){2-3}\cmidrule(lr){4-5}
~ & \#Params~$^2$ & GFLOPs~$^2$ & ~\#Params~ & ~GFLOPs~ & ~ & ~ & ~ \\
\noalign{\smallskip}
\hline
\noalign{\smallskip}
\multicolumn{1}{l|}{Backbone: SG-Former-T} & 7.84M & \multicolumn{1}{c|}{46.1} & 6.79M & \multicolumn{1}{c|}{45.7} & - & - & - \\
\multicolumn{1}{l|}{~~~Patch embedding} & 0.08M & \multicolumn{1}{c|}{12.6} & 0.08M & \multicolumn{1}{c|}{12.6} & - & $W/4 \times H/4$ & \phantom{0}64 \\
\multicolumn{1}{l|}{~~~Stage 1} & 0.38M & \multicolumn{1}{c|}{\phantom{0}6.9} & 0.12M & \multicolumn{1}{c|}{\phantom{0}6.9} & 2 & $W/4 \times H/4$ & \phantom{0}64 \\
\multicolumn{1}{l|}{~~~Down-sampling embedding 1} & 0.03M & \multicolumn{1}{c|}{\phantom{0}0.8} & 0.03M & \multicolumn{1}{c|}{\phantom{0}0.8} & - & $W/8 \times H/8$ & 128 \\
\multicolumn{1}{l|}{~~~Stage 2} & 1.41M & \multicolumn{1}{c|}{13.5} & 0.88M & \multicolumn{1}{c|}{13.4} & 4 & $W/8 \times H/8$ & 128 \\
\multicolumn{1}{l|}{~~~Down-sampling embedding 2} & 0.10M & \multicolumn{1}{c|}{\phantom{0}0.8} & 0.10M & \multicolumn{1}{c|}{\phantom{0}0.8} & - & $W/16 \times H/16$ & 256 \\
\multicolumn{1}{l|}{~~~Stage 3} & 2.00M & \multicolumn{1}{c|}{\phantom{0}7.0} & 1.74M & \multicolumn{1}{c|}{\phantom{0}6.8} & 2 & $W/16 \times H/16$ & 256 \\
\multicolumn{1}{l|}{~~~Down-sampling embedding 3} & 0.40M & \multicolumn{1}{c|}{\phantom{0}0.8} & 0.40M & \multicolumn{1}{c|}{\phantom{0}0.8} & - & $W/32 \times H/32$ & 512 \\
\multicolumn{1}{l|}{~~~Stage 4} & 3.44M & \multicolumn{1}{c|}{\phantom{0}3.5} & 3.44M & \multicolumn{1}{c|}{\phantom{0}3.5} & 1 & $W/32 \times H/32$ & 512 \\
\hline\noalign{\smallskip}
\multicolumn{8}{l}{\scalebox{1.00}{~$^1$ ``With global'' and ``Without global'' denote SG-Former-T with and without global attention, respectively.}} \\
\multicolumn{8}{l}{\scalebox{1.00}{~$^2$ Computational costs were calculated using~\cite{MMCV_MISC_2018}.}} \\
\multicolumn{8}{l}{\scalebox{1.00}{~$^3$ ``Size'' denotes the spatial dimensions of feature maps relative to the input image of size $W \times H$.}} \\
\end{tabular}
}
\vspace{6mm}
\caption{Comparison of computational costs and the average precision (AP) $\uparrow$ on the LOAF test set}
\label{table-comparison-of-computational-costs-and-the-average-precision-on-the-loaf-test-set}
\centering
\scalebox{0.88}{
\begin{tabular}{ccc|ccc|ccc|cc}
\hline\noalign{\smallskip}
Method & \#Params~$^1$ & GFLOPs~$^1$ & mAP & AP$_{50}$ & AP$_{75}$ & AP$_n$ & AP$_m$ & AP$_f$ & AP$_{seen}$ & AP$_{unseen}$ \\
\noalign{\smallskip}
\hline
\noalign{\smallskip}
Seidel~\etal{}~\cite{Seidel_VISAPP_2019} & \phantom{0}59.22M & 2518.6 & 20.2 & 58.2 & \phantom{0}7.1 & 32.2 & 28.3 & \phantom{0}3.9 & 22.4 & 18.9 \\
Li~\etal{}~\cite{Li_AVSS_2019} & \phantom{0}61.98M & 1072.4 & 27.2 & 65.2 & 21.3 & 47.2 & 24.8 & \phantom{0}1.3 & 31.8 & 27.6 \\
~~Tamura~\etal{}~\cite{Tamura_WACV_2019}~~ & \phantom{0}59.22M & \phantom{0}123.1 & 34.2 & 72.8 & 28.7 & 53.7 & 37.3 & \phantom{0}6.5 & 39.5 & 33.2 \\
RAPiD~\cite{Duan_CVPRW_2020} & \phantom{0}61.52M & \phantom{0}198.5 & 39.2 & 77.9 & 35.4 & 54.8 & 40.1 & \phantom{0}7.9 & 44.2 & 37.3 \\
OARPD~\cite{Qiao_MTA_2024} & \phantom{0}26.42M & \phantom{00}81.2 & 45.8 & 79.4 & 45.8 & - & - & - & - & - \\
Yang~\etal{}~\cite{Yang_ICCV_2023}~$^2$ & \phantom{0}48.23M & \phantom{0}280.5 & 46.2 & 81.1 & 47.3 & \textbf{66.1} & 53.5 & 12.6 & 49.3 & 44.9 \\
\noalign{\smallskip}
\hline\hline
\noalign{\smallskip}
\multicolumn{3}{l|}{Backbone: SG-Former-S ($N$[2, 4, 16, 1])} \\
\multicolumn{1}{l}{~~~Ours (W3072~$^3$)} & \phantom{0}42.00M & \phantom{0}624.0 & 54.5 & 87.2 & 61.1 & 62.8 & 60.5 & 44.6 & 56.2 & 53.6 \\
\multicolumn{1}{l}{~~~Ours (W3072, w/o PDAT~$^4$)} & \phantom{0}42.00M & \phantom{0}624.0 & 49.8 & 84.7 & 52.7 & 59.3 & 55.3 & 39.7 & 51.0 & 48.9 \\
\multicolumn{1}{l}{~~~Ours (W2560)} & \phantom{0}42.00M & \phantom{0}434.0 & 51.2 & 85.1 & 55.6 & 61.9 & 56.4 & 39.3 & 53.8 & 49.7 \\
\multicolumn{1}{l}{~~~Ours (W2048)} & \phantom{0}42.00M & \phantom{0}278.6 & 47.0 & 81.3 & 49.2 & 58.3 & 52.0 & 35.5 & 48.5 & 46.0 \\
\multicolumn{1}{l}{~~~Ours (W2048, w/o global~$^4$)} & \phantom{0}39.11M & \phantom{0}276.3 & 44.9 & 78.8 & 46.2 & 56.3 & 48.8 & 30.9 & 47.0 & 43.8 \\
\noalign{\smallskip}
\hline
\noalign{\smallskip}
\multicolumn{3}{l|}{Backbone: SG-Former-T ($N$[2, 4, 2, 1])} \\
\multicolumn{1}{l}{~~~Ours (W2048)} & \phantom{0}28.01M & \phantom{0}229.4 & 47.3 & 82.0 & 49.4 & 56.4 & 52.8 & 36.5 & 48.1 & 46.9 \\
\multicolumn{1}{l}{~~~Ours (W2048, w/o global)~$^2$} & \phantom{0}\textbf{26.96M} & \phantom{0}\textbf{229.0} & \textbf{48.0} & \textbf{82.5} & \textbf{50.7} & 58.0 & \textbf{54.0} & \textbf{37.1} & \textbf{49.4} & \textbf{47.1} \\
\hline\noalign{\smallskip}
\multicolumn{11}{l}{\scalebox{1.00}{~$^1$ Computational costs were calculated using~\cite{MMCV_MISC_2018}.}} \\
\multicolumn{11}{l}{\scalebox{1.00}{~$^2$ Boldface indicates the better result between Yang~\etal{}~\cite{Yang_ICCV_2023} and Ours (W2048, w/o global) using SG-Former-T.}} \\
\multicolumn{11}{l}{\scalebox{1.00}{~$^3$ W3072, W2560, and W2048 denote panoramic image widths of 3072, 2560, and 2048 pixels, respectively.}} \\
\multicolumn{11}{l}{\scalebox{1.00}{~$^4$ ``w/o PDAT'' and ``w/o global'' denote our method without PDAT and global attention, respectively.}} \\
\end{tabular}
}
\end{minipage}
\end{table*}

\section{Computational costs}
\label{sec-computational-costs}
We describe the computational costs of our proposed method. We also introduce a lightweight variant of SG-Former that outperforms Yang~\etal{}'s method~\cite{Yang_ICCV_2023} in terms of both detection accuracy and computational costs.

\subsection{Analysis of SG-Former computational costs}
\label{sec-analysis-of-sg-former-computational-costs}
We measured the computational costs of our method using SG-Former-S, which has the smallest backbone size among the official S, M, and B variants in~\cite{Ren_ICCV_2023}. \tref{table-Computational-costs-of-SG-Former-S-and-the-detector-head} shows the details of the computational costs of SG-Former-S and DAB-DETR using panoramic image widths of 3072, 2560, and 2048 pixels. We observed that stage 3 has the largest number of parameters and GFLOPs among SG-Former submodules; \ie{}, stage 3 occupied 73\% of the parameters in SG-Former-S. The number of parameters of SG-Former-S exceeds that of the DAB-DETR detector head, indicating substantially higher computational costs in the backbone.

On the basis of the observation above, we introduce a lightweight variant, called SG-Former-T. The block counts of SG-Former-T in stages 1--4 are 2, 4, 2, and 1, denoted as $N[2, 4, 2, 1]$; in contrast, SG-Former-S has $N[2, 4, 16, 1]$. To further reduce computational costs, we explored the attention mechanism of SG-Former. We found that the hybrid-scale self-attention produces global and local attention maps independently. The global attention accounts for the entire feature map, while the local attention uses window partitioning. This global attention does not effectively extract information from panoramic images because small persons dominate the number of instances, as shown in \fref{fig-bbox-distribution}(b) (main paper). \tref{table-Computational-costs-of-SG-Former-T-using-panoramic-image-width-of-2048-pixels} shows the computational costs of SG-Former-T with or without the global attention. In the panoramic image with a width of 2048 pixels, SG-Former-T with the global attention has 7.84M parameters and 46.1 GFLOPs, while SG-Former-T without the global attention has 6.79M parameters and 45.7 GFLOPs. Therefore, SG-Former-T has superior computational efficiency to SG-Former-S, with 21.83M parameters and 95.3 GFLOPs.

\subsection{Comparison of computational costs}
\label{sec-comparison-of-computational-costs}
We compared the computational costs of conventional methods with those of our method. \tref{table-comparison-of-computational-costs-and-the-average-precision-on-the-loaf-test-set} shows the computational costs and the average precision (AP) on the LOAF test set. We used the mean average precision (mAP) following the COCO evaluation ~\cite{Lin_ECCV_2014}. We used the same subscripts, such as $n$, $m$, and $f$, as described in \sref{sec-person-detection-on-loaf} (main paper). Yang~\etal{}'s method~\cite{Yang_ICCV_2023}, the existing state-of-the-art method with an mAP of 46.2, has 48.23M parameters and 280.5 GFLOPs. Our method using a 3072-pixel image width has 42.00M parameters and 624.0 GFLOPs, achieving the highest mAP of 54.5, as listed in \tref{table-comparison-of-computational-costs-and-the-average-precision-on-the-loaf-test-set}, although it has a large computational cost. To address the computational cost problem, our lightweight variant achieved a higher mAP of 48.0 than that of Yang~\etal{}'s method~\cite{Yang_ICCV_2023}. This lightweight variant uses a 2048-pixel image width and SG-Former-T without global attention, which has 26.96M parameters and 229.0 GFLOPs, as shown in the bottom row of \tref{table-comparison-of-computational-costs-and-the-average-precision-on-the-loaf-test-set}. Furthermore, the AP$_f$ value obtained using this lightweight variant was remarkably larger than that of Yang~\etal{}'s method~\cite{Yang_ICCV_2023} by 24.5.

\begin{table*}[t]
\caption{Comparison of scale factors in our method for person detection on the LOAF validation (val) and test sets}
\label{table-comparison-of-scale-factors-in-our-method-for-person-detection-on-the-loaf-validation-and-test-sets}
\centering
\scalebox{0.77}{
\begin{tabular}{cccc|ccc|cc||ccc|ccc|cc}  
\hline\noalign{\smallskip}
\multirow{2}{*}{Method} & \multicolumn{8}{c||}{$\texttt{val}$} & \multicolumn{8}{c}{$\texttt{test}$} \\
\cmidrule(lr){2-9}\cmidrule(lr){10-17}
~ & mAP & AP$_{50}$ & AP$_{75}$ & AP$_n$ & AP$_m$ & AP$_f$ & AP$_{seen}$ & AP$_{unseen}$  & mAP & AP$_{50}$ & AP$_{75}$ & AP$_n$ & AP$_m$ & AP$_f$ & AP$_{seen}$ & AP$_{unseen}$ \\
\noalign{\smallskip}
\hline\noalign{\smallskip}
Ours ($\alpha = 1$) & 47.7 & 82.2 & 49.9 & 63.1 & 53.3 & 37.0 & 51.3 & 46.1 & 49.8 & 84.7 & 52.7 & 59.3 & 55.3 & 39.7 & 51.0 & 48.9 \\
Ours ($\alpha = 2$) & \textbf{52.4} & \textbf{84.6} & \textbf{57.6} & 66.3 & \textbf{57.9} & \textbf{42.9} & \textbf{56.3} & \textbf{50.7} & \textbf{54.5} & \textbf{87.2} & \textbf{61.1} & \textbf{62.8} & \textbf{60.5} & \textbf{44.7} & \textbf{56.2} & \textbf{53.6} \\
Ours ($\alpha = 3$) & 51.8 & 84.1 & 56.4 & \textbf{66.7} & 57.3 & 41.9 & 55.7 & 50.0 & 53.9 & 86.5 & 60.1 & 61.9 & \textbf{60.5} & 44.6 & 56.1 & 52.6 \\
\hline\noalign{\smallskip}
\end{tabular}
}
\end{table*}
\begin{table*}[t]
\caption{Comparison of properties of the overhead fisheye image datasets}
\label{table-comparison-of-properties-of-the-overhead-fisheye-image}
\centering
\scalebox{0.90}{
\begin{tabular}{ccccccc}
\hline\noalign{\smallskip}
\multicolumn{2}{c}{\multirow{2}{*}{Dataset}} & \multirow{2}{*}{Availability} & \multirow{2}{*}{Max resolution} & \multirow{2}{*}{Bounding box} & \multicolumn{2}{c}{Number of images} \\
\cmidrule(lr){6-7}
~ & ~ & ~ & ~ & ~ & Train~/~Validation~/~Test & All \\
\hline\noalign{\smallskip}
HABBOF~\cite{Li_AVSS_2019} & AVSS'19 & \checkmark\phantom{~$^1$} & 2048 $\times$ 2048 & Human-aligned & - & \phantom{0,}5837 \\
MW-R~\cite{Duan_CVPRW_2020} & CVPRW'20 & ~ & 1488 $\times$ 1488 & Human-aligned & - & \phantom{0,}8752 \\
CEPDOF~\cite{Duan_CVPRW_2020} & CVPRW'20 & \checkmark\phantom{~$^1$} & 2048 $\times$ 2048 & Human-aligned & - & 25,504 \\
WEPDTOF~\cite{Tezcan_WACV_2022} & WACV'22 & \checkmark\phantom{~$^1$} & 2592 $\times$ 1944 & Human-aligned & - & 10,544 \\
LOAF~\cite{Yang_ICCV_2023} & ICCV'23 &\checkmark~$^1$ & 2048 $\times$ 2048 & Radius-aligned & 29,569~/~4600~/~8773 & 42,942 \\
\hline\noalign{\smallskip}
\multicolumn{7}{l}{~$^1$ Images with the resolution of 2952 $\times$ 2952 are not publicly available.} \\
\end{tabular}
}
\end{table*}
We analyzed the effects of using SG-Former-T and removing global attention. These modifications were introduced to reduce computational costs, as discussed in \sref{sec-analysis-of-sg-former-computational-costs}. In the case of a 2048-pixel image width, the AP value using SG-Former-T was slightly larger than that using SG-Former-S by 0.3, indicating that the large number of parameters in stage 3 of SG-Former-S is redundant. Similarly, the mAP value of SG-Former-T without global attention was greater than that of SG-Former-T with global attention by 0.7, indicating the redundancy of the global attention. Therefore, these modifications demonstrate the effectiveness of our lightweight variant in addressing the computational cost problem.

\section{Experimental details}
\label{sec-experimental-details}
We provide the experimental details, focusing on the number of regions, scale factors, and evaluation results of the CEPDOF and WEPDTOF datasets. Unless otherwise specified, we used the same experimental settings, \eg{}, using SG-Former-S and panoramic images with a 3072-pixel width, as described in \sref{sec-parameter-settings} (main paper).

\subsection{Details of the number of regions}
We describe the relation between the number of regions $K = 5$ and PDAT tiling in our experiments, as described in \sref{sec-parameter-settings} (main paper). \fref{fig-our-tokenization} (main paper) illustrates the tiling using $K = 5$. The length of the horizontally 32 repeated unit figures corresponds to the width of the feature maps. The smallest tile width in region 1 is 1/16 (1/$2^{K-1}$) of the width of the unit figure; thus, 512 ($= 32 \times 16$) tiles in region 1 align horizontally. The width of the feature maps of the patch embeddings is $W/4$, where $W$ is the panoramic image width, as indicated in \tref{table-Computational-costs-of-SG-Former-S-and-the-detector-head}. Therefore, PDAT using $K = 5$ is suitable for panoramic images with a width of 2048 pixels ($= 512 \times 4$). We also employed $K = 5$ for panoramic images with a width of 3072 pixels. To handle floating-point tile boundaries, we rounded these boundaries.

\subsection{Comparison of scale factors}
To analyze the dependence of the scale factors $\alpha$, we trained our networks with different values of $\alpha$. This scale factor multiplies the significance map $\mathbf{S}$ of SG-Former, as described in \sref{sec-network-architecture} (main paper). \tref{table-comparison-of-scale-factors-in-our-method-for-person-detection-on-the-loaf-validation-and-test-sets} shows the results of our method using different scale factors on the LOAF validation (val) and test sets. The first row represents our method without PDAT because $\alpha = 1$ means that the significance maps are not changed. The second row demonstrates the effectiveness of PDAT with $\alpha = 2$, corresponding to the default value in our experiments. The third row indicates the case of a larger scale factor: $\alpha = 3$, and the mAP values of our method with $\alpha = 3$ were lower than those with $\alpha = 2$ by 0.6 on both the validation and test sets. This degradation suggests that too large scale factors cause the imbalance of the PDAT leveraging process. Therefore, the scale factor $\alpha = 2$ led to the best performance among 1, 2, and 3.

\subsection{Datasets}
\begin{table*}[t]
\caption{Person detection results including Yang~\etal{}'s report~\cite{Yang_ICCV_2023} on the CEPDOF dataset}
\label{table-person-detection-result-on-cepdof-full}
\centering
\scalebox{0.90}{
\begin{tabular}{ccccc|ccc|ccc}  
\hline\noalign{\smallskip}
\multirow{2}{*}{Method} & \multicolumn{4}{c}{Training dataset} & \multicolumn{6}{c}{Metric} \\
\cmidrule(lr){2-5}\cmidrule(lr){6-11}
~ & COCO~$^1$ & HABBOF & MW-R & WEPDTOF & mAP & AP$_{50}$ & AP$_{75}$ & Precision & Recall & F1-score \\
\noalign{\smallskip}
\hline
\hline\noalign{\smallskip}
\multicolumn{11}{c}{Results~\cite{Yang_ICCV_2023} using an unavailable dataset (MW-R)} \\
\hline\noalign{\smallskip}
Seidel~\etal{}~\cite{Seidel_VISAPP_2019} & Pretrain & ~ & ~ & ~ & 20.9 & 50.6 & 10.2 & 80.6 & 39.5 & 53.0 \\
Li~\etal{}~\cite{Li_AVSS_2019} & Pretrain & ~ & ~ & ~ & 34.2 & 75.7 & 28.6 & 86.3 & 65.4 & 74.4 \\
Tamura~\etal{}~\cite{Tamura_WACV_2019} & Pretrain & \checkmark & \checkmark & \checkmark & 29.3 & 61.0 & 23.4 & 88.8 & 51.2 & 65.0 \\
RAPiD~\cite{Duan_CVPRW_2020} & Pretrain & \checkmark & \checkmark & \checkmark & 39.3 & 85.4 & 26.0 & 89.2 & 78.7 & 83.6 \\
Yang~\etal{}~\cite{Yang_ICCV_2023} & Pretrain & \checkmark & \checkmark & \checkmark & \textbf{47.2} & \textbf{88.5} & \textbf{38.2} & \textbf{90.3} & \textbf{87.6} & \textbf{88.9} \\
\hline\hline\noalign{\smallskip}
\multicolumn{11}{c}{Results using available datasets} \\
\hline\noalign{\smallskip}
Seidel~\etal{}~\cite{Seidel_VISAPP_2019} & Pretrain & ~ & ~ & ~ & 20.9 & 50.6 & 10.2 & 80.6 & 39.5 & 53.0 \\
Li~\etal{}~\cite{Li_AVSS_2019} & Pretrain & ~ & ~ & ~ & 34.2 & 75.7 & 28.6 & 86.3 & 65.4 & 74.4 \\
Tamura~\etal{}~\cite{Tamura_WACV_2019}~$^2$ & Pretrain & \checkmark & ~ & \checkmark & 28.8 & 61.6 & 23.0 & 73.1 & 65.8 & 69.3 \\
RAPiD~\cite{Duan_CVPRW_2020}~$^2$ & Pretrain & \checkmark & ~ & \checkmark & 38.3 & 78.6 & 30.8 & \textbf{91.8} & 80.3 & 85.7 \\
OARPD~\cite{Qiao_MTA_2024}~$^2$ & Pretrain & \checkmark & ~ & \checkmark & 11.4 & 28.9 & \phantom{0}6.6 & 80.5 & 31.3 & 45.1 \\
Yang~\etal{}~\cite{Yang_ICCV_2023}~$^2$ & Pretrain & \checkmark & ~ & \checkmark & 39.7 & 78.3 & 37.1 & 82.9 & 72.6 & 77.4 \\
\hline\noalign{\smallskip}
Ours & Pretrain & \checkmark & ~ & \checkmark & \textbf{42.8} & \textbf{85.6} & \textbf{38.7} & \textbf{91.8} & \textbf{81.8} & \textbf{86.5} \\
\hline\noalign{\smallskip}
\multicolumn{11}{l}{~$^1$ ``Pretrain'' represents that COCO is used for pretraining.} \\
\multicolumn{11}{l}{~$^2$ We trained networks for evaluation.}
\end{tabular}
}
\end{table*}
\tref{table-comparison-of-properties-of-the-overhead-fisheye-image} shows the properties of overhead fisheye image datasets with bounding boxes. LOAF provides the official dataset splitting: training, validation, and test sets. In contrast, the other datasets do not have them. Therefore, prior works~\cite{Tezcan_WACV_2022, Yang_ICCV_2023} used MW-R, CEPDOF, and WEPDTOF for the training datasets. However, the MW-R dataset is not publicly available. Furthermore, unlike the numerous publicly available perspective image datasets, overhead fisheye datasets are limited, because overhead fisheye cameras need to be attached to the ceilings or capturing apparatus. Following \cite{Yang_ICCV_2023}, we converted the human-aligned bounding boxes into the radius-aligned ones because the angles of human-aligned bounding boxes are ambiguous; \ie, the angles depend on the annotators.

\subsection{Person detection on CEPDOF and WEPDTOF}

\subsubsection{Details of the training}
As described in \sref{sec-person-detection-on-cepdof} (main paper), for person detection on CEPDOF, we trained the method of Tamura~\etal{}~\cite{Tamura_WACV_2019}, along with RAPiD~\cite{Duan_CVPRW_2020}, OARPD~\cite{Qiao_MTA_2024}, Yang~\etal{}~\cite{Yang_ICCV_2023}, and our method, using HABBOF~\cite{Li_AVSS_2019} and WEPDTOF~\cite{Tezcan_WACV_2022} after pretraining these methods on COCO~\cite{Lin_ECCV_2014}, although MW-R~\cite{Duan_CVPRW_2020} was not used for the training. In the evaluation of WEPDTOF, we used HABBOF and CEPDOF for the training. We found that the lack of the MW-R dataset, consisting of 8752 images, led to overfitting when we used the training procedures provided by the corresponding papers. In addition, CEPDOF and WEPDTOF do not have the official validation sets, as shown in \tref{table-comparison-of-properties-of-the-overhead-fisheye-image}. Therefore, we reported each maximum mAP within the training in both conventional methods and our method.

\subsubsection{Results on CEPDOF}
We compared the results reported by Yang~\etal{}~\cite{Yang_ICCV_2023} using MW-R and our experiments. The mAP values achieved using the MW-R dataset for the training were higher than those obtained without MW-R, as listed in \tref{table-person-detection-result-on-cepdof-full}. The mAP for Yang~\etal{}'s method~\cite{Yang_ICCV_2023} using MW-R was larger than that without MW-R by 4.4. The mAP in our method was comparable to Yang~\etal{}'s method~\cite{Yang_ICCV_2023} using MW-R. When MW-R was not used for the training, the HABBOF and WEPDTOF datasets provided diverse images because WEPDTOF consists of 16 videos, including various scenes.

\subsubsection{Results on WEPDTOF}
Similar to the CEPDOF discussion described above, we compared Yang~\etal{}'s report~\cite{Yang_ICCV_2023} and our experiments. In the evaluation of WEPDTOF, we trained networks on the HABBOF and CEPDOF datasets. The combination of these datasets lacks image diversity, yielding the degradation of the detection performance. The mAP values obtained using methods of Tamura~\etal{}~\cite{Tamura_WACV_2019}, RAPiD~\cite{Duan_CVPRW_2020}, OARPD~\cite{Qiao_MTA_2024}, and Yang~\etal{}~\cite{Yang_ICCV_2023} decreased notably, as evidenced by the drop of mAP from 45.8 (with MW-R) to 20.5 (without MW-R) in Yang~\etal{}'s method~\cite{Yang_ICCV_2023}. In contrast, the mAP value obtained using our method was 37.5, outperforming conventional methods without MW-R for the training. This result suggests that our method accurately detects bounding boxes on panoramic images without person rotation, unlike conventional methods that detect on fisheye images.
\begin{table*}[t]
\caption{Person detection results including Yang~\etal{}'s report~\cite{Yang_ICCV_2023} on the WEPDTOF dataset}
\label{table-person-detection-result-on-wepdtof}
\centering
\scalebox{0.90}{
\begin{tabular}{ccccc|ccc|ccc}  
\hline\noalign{\smallskip}
\multirow{2}{*}{Method} & \multicolumn{4}{c}{Training dataset} & \multicolumn{6}{c}{Metric} \\
\cmidrule(lr){2-5}\cmidrule(lr){6-11}
~ & COCO~$^1$ & HABBOF & MW-R & CEPDOF & mAP & AP$_{50}$ & AP$_{75}$ & Precision & Recall & F1-score \\
\noalign{\smallskip}
\hline
\hline\noalign{\smallskip}
\multicolumn{11}{c}{Results~\cite{Yang_ICCV_2023} using an unavailable dataset (MW-R)} \\
\hline\noalign{\smallskip}
Seidel~\etal{}~\cite{Seidel_VISAPP_2019} & Pretrain & ~ & ~ & ~ & 16.1 & 39.4 & \phantom{0}9.0 & 70.9 & 38.6 & 50.0 \\
Li~\etal{}~\cite{Li_AVSS_2019} & Pretrain & ~ & ~ & ~ & 25.2 & 69.9 & 30.2 & 81.4 & 64.5 & 72.0 \\
Tamura~\etal{}~\cite{Tamura_WACV_2019} & Pretrain & \checkmark & \checkmark & \checkmark & 28.8 & 59.8 & 24.2 & 77.0 & 52.4 & 62.4 \\
RAPiD~\cite{Duan_CVPRW_2020} & Pretrain & \checkmark & \checkmark & \checkmark & 37.7 & 72.0 & 26.8 & 73.3 & 67.8 & 70.4 \\
Yang~\etal{}~\cite{Yang_ICCV_2023} & Pretrain & \checkmark & \checkmark & \checkmark & \textbf{45.8} & \textbf{85.6} & \textbf{36.6} & \textbf{84.9} & \textbf{74.6} & \textbf{79.8} \\
\hline\hline\noalign{\smallskip}
\multicolumn{11}{c}{Results using available datasets} \\
\hline\noalign{\smallskip}
Seidel~\etal{}~\cite{Seidel_VISAPP_2019} & Pretrain & ~ & ~ & ~ & 16.1 & 39.4 & \phantom{0}9.0 & 70.9 & 38.6 & 50.0 \\
Li~\etal{}~\cite{Li_AVSS_2019} & Pretrain & ~ & ~ & ~ & 25.2 & 69.9 & 30.2 & 81.4 & 64.5 & 72.0 \\
Tamura~\etal{}~\cite{Tamura_WACV_2019}~$^2$ & Pretrain & \checkmark & ~ & \checkmark & 12.8 & 27.2 & 10.9 & 53.0 & 34.3 & 41.6 \\
RAPiD~\cite{Duan_CVPRW_2020}~$^2$ & Pretrain & \checkmark & ~ & \checkmark & 25.0 & 54.0 & 17.4 & \textbf{83.7} & 57.6 & 68.2 \\
OARPD~\cite{Qiao_MTA_2024}~$^2$ & Pretrain & \checkmark & ~ & \checkmark & \phantom{0}4.1 & 10.0 & \phantom{0}2.2 & 28.8 & 19.0 & 22.9 \\
Yang~\etal{}~\cite{Yang_ICCV_2023}~$^2$ & Pretrain & \checkmark & ~ & \checkmark & 20.5 & 45.3 & 16.3 & 66.8 & 43.0 & 52.3 \\
\hline\noalign{\smallskip}
Ours & Pretrain & \checkmark & ~ & \checkmark & \textbf{37.5} & \textbf{74.9} & \textbf{34.8} & 81.7 & \textbf{69.0} & \textbf{74.8} \\
\hline\noalign{\smallskip}
\multicolumn{11}{l}{~$^1$ ``Pretrain'' represents that COCO is used for pretraining.} \\
\multicolumn{11}{l}{~$^2$ We trained networks for evaluation.}
\end{tabular}
}
\end{table*}

\section{Extended related work}
\label{sec-extended-related-work}
Due to the space limitations of the main paper, we provide an extended review of related work, focusing on object detection and panoramic-image conversion.

\textbf{Object detection.}
Object detection is a fundamental computer vision task that estimates the positions and class labels of objects in images. The person class is one of the most commonly targeted in object detection because of the wide range of applications, such as visual surveillance. In this paper, we focus on single-class person detection, \ie{}, detecting only person instances. Object detection methods can be classified into two categories based on the architecture: convolutional neural network (CNN)-based and transformer-based methods. The CNN-based methods use convolutional backbones to extract multi-scale feature maps to handle scale variation, \eg{}, Faster R-CNN~\cite{Ren_NIPS_2015}, SSD~\cite{Liu_ECCV_2016}, and YOLOv10~\cite{Wang_NIPS_2024}. Transformer-based methods achieve end-to-end training with attention mechanisms, yielding accurate detection, \eg{}, DAB-DETR~\cite{Liu_ICLR_2022}, Mr. DETR~\cite{Zhang_2025_CVPR} and DEIM~\cite{Huang_2025_CVPR}. Although both CNN-based and transformer-based methods achieve strong performance on perspective images, their performance substantially degrades on overhead fisheye images due to challenges of person rotation and the presence of small persons.

\textbf{Panoramic-image conversion.}
When the camera parameters are given, the panoramic images, fisheye images, and spherical representations are all convertible. These conversion methods can be applied in various tasks using fisheye and panoramic images. To alleviate fisheye distortion, Plaut~\etal{}~\cite{Plaut_CVPRW_2021} converted fisheye images into equirectangular images to perform 3D object detection. In deep single image camera calibration, equirectangular images were converted into fisheye images for dataset generation~\cite{Lopez_CVPR_2019, Wakai_CVPR_2024}. Wang~\etal{}~\cite{Wang_CVPR_2021} used spherical representations taken from equirectangular images to perform 360$^\circ$ room layout estimation. As described in the discussion of overhead-fisheye-image person detection in \sref{sec-related-work} (main paper), conventional methods do not convert the fisheye images into panoramic images because person detection in panoramic images remains an open challenge.

\section{Novelty}
\label{sec-novelty}
To describe the novelty of the paper, we reiterate our major contributions:
\begin{itemize}
\setlength{\itemsep}{3mm}
\item We propose a transformer-based method for person detection and localization in overhead fisheye images that achieves higher accuracy than conventional methods by balancing significance areas of the feature maps.

\item We introduce the PDAT method with optimal divisions based on self-similar figures to address a person's height using the vertical panoramic-image coordinates.

\item We demonstrate the effectiveness of our proposed method by applying it to the challenging LOAF dataset. The average precision for distant person positions when using our method on the LOAF test set is substantially greater than that of the existing state-of-the-art method by 32.0.
\end{itemize}
In the remainder of this section, we explain the novelty of the paper in terms of our contributions. 

\textbf{Transformer-based method.}
As the first contribution, our transformer-based method achieved accurate person detection in overhead fisheye images. Conventional methods do not convert input fisheye images. In contrast, our method converts an input fisheye image into a panoramic image using remapping. This conversion enables our network to detect persons accurately because panoramic images do not contain rotated persons, unlike fisheye images. Additionally, panoramic images, except for beneath cameras, are less distorted than fisheye images. Furthermore, our transformer-based architecture effectively leverages attention mechanisms to preserve the significance areas of smaller persons. While conventional detection methods tend to favor larger persons in their attention maps, our transformer-based method, combined with PDAT, achieves accurate person detection and localization. Therefore, our transformer-based detector can detect various persons in overhead fisheye images.

\textbf{Panoramic distortion-aware tokenization.}
As the second contribution, PDAT divides panoramic features using self-similar figures and leverages the maximum significance values in each tile to preserve the significance areas of smaller persons. This approach enables our network to detect small persons effectively, even in the peripheral regions of fisheye images; \ie{}, our method can detect persons far from cameras. Our method detected tiny persons that the conventional method~\cite{Yang_ICCV_2023} did not detect.

\textbf{State-of-the-art performance.}
As the third contribution, our method outperformed the existing state-of-the-art method by substantial margins in different metrics, thus improving the performance in the various applications. Note that the proposed method is particularly effective for applications in larger spaces because our method can detect persons at over 20 m from overhead fisheye cameras, as demonstrated in our experiments.

Our contributions advance the practical use of overhead fisheye cameras, which provide wider fields of view than narrow-view cameras and fewer occlusions than handheld cameras due to their ceiling-mounted positions. The wide coverage of these cameras enables efficient monitoring of large areas such as shops, factories, and outdoor spaces using only a few cameras. Our method accurately detects both rotated and small persons in overhead fisheye images, and enables various applications as follows:
\vspace{2mm}
\begin{enumerate}
\item \textbf{Security.} In stations, airports, and public facilities, our method enables efficient visual surveillance to detect suspicious behavior. The wide field of view of overhead fisheye cameras allows comprehensive monitoring with fewer cameras, reducing operational costs.

\item \textbf{Smart city and buildings.} Our method supports smart city and building management through real-time monitoring of crowd density and people counting. These capabilities enable efficient space utilization and improved building operations.

\item \textbf{Manufacturing.} In factories and warehouses, our method enhances safety management and workflow optimization by analyzing worker movement patterns. The ceiling-mounted cameras provide clear visibility even in environments with tall shelves and machinery.

\item \textbf{Medical and welfare.} Our method provides contactless monitoring in healthcare facilities, offering an advantage over burdensome wearable devices for patients and elderly residents.

\item \textbf{Marketing.} Our method provides comprehensive monitoring of customer behavior across entire floor spaces, enabling detailed analysis of traffic patterns, unlike conventional approaches that capture only partial areas.
\end{enumerate}

\vspace{3mm}
Our contributions can extend beyond person detection and localization, offering potential benefits to various computer vision tasks. For example, in 3D pose estimation and action recognition, overhead fisheye images provide advantages in measuring precise positions and movements of people because these cameras have a wide field of view and reduce occlusions. Furthermore, these advantages can enhance human--machine interaction applications by enabling spatial understanding of both people and objects within large spaces. Therefore, our approach provides fundamental techniques that benefit both theoretical research and real-world applications in computer vision.

\section{Broader impact}
\label{sec-broader-impact}
Our person detection method can be used in a wide range of applications, including visual surveillance in public facilities and safety management in factories. Additionally, our method can detect small persons at locations far away from the cameras in these applications. Although our method offers these advantages, we acknowledge that all surveillance technologies have the potential to cause privacy concerns. We are committed to promoting responsible use of our proposed method while also addressing these privacy concerns.

\section{Qualitative evaluation}
\label{sec-qualitative-evaluation}
\begin{figure*}[t]
\centering
\includegraphics[width=0.85\hsize]{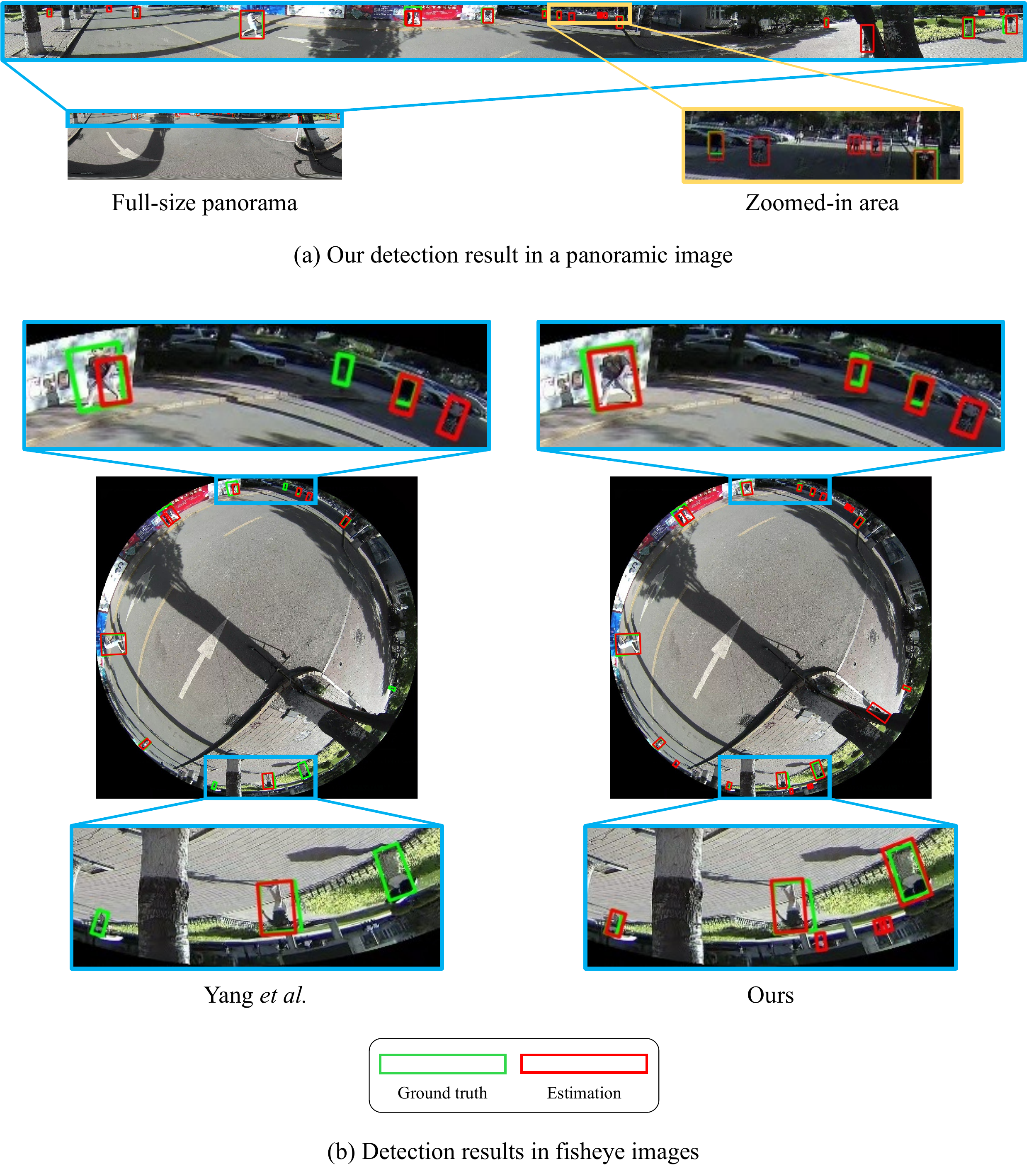}
\caption{Qualitative results achieved on the LOAF test set. Green and red rectangles indicate the ground-truth and estimated bounding boxes, respectively. All methods show detection results with confidence scores greater than 0.3. (a) Our detection results in a panoramic image showing a full-size view and its zoomed-in area. (b) The detection results of Yang~\etal{}~\cite{Yang_ICCV_2023} on the left side and our method on the right side.}
\label{fig-qualitative-result}
\end{figure*}

\begin{figure*}[t]
\centering
\includegraphics[width=0.75\hsize]{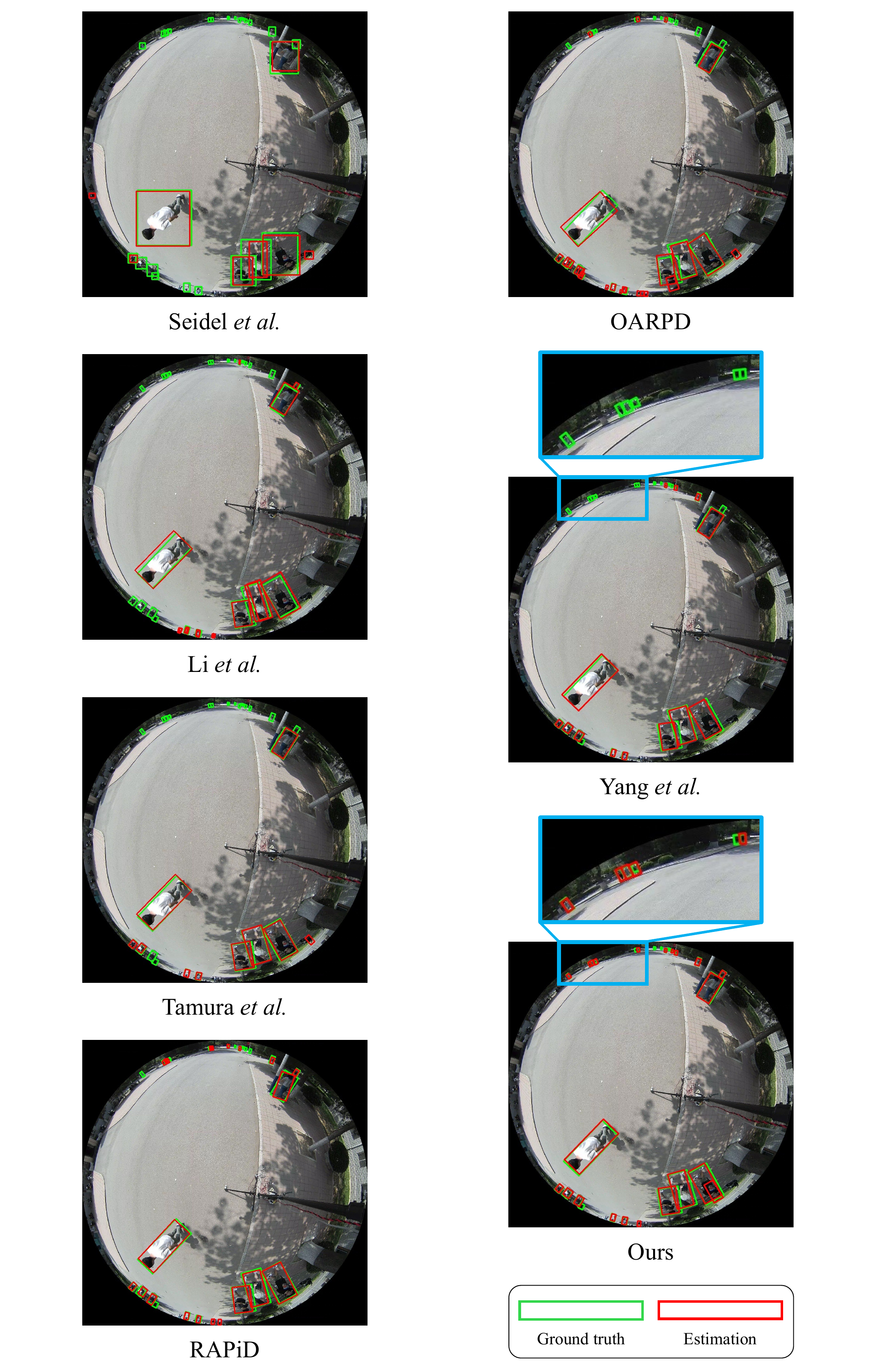}
\caption{Qualitative results achieved on the LOAF test set. The first column from top to bottom shows the results of methods of Seidel~\etal{}\cite{Seidel_VISAPP_2019}, Li~\etal{}~\cite{Li_AVSS_2019}, Tamura~\etal{}~\cite{Tamura_WACV_2019}, and RAPiD~\cite{Duan_CVPRW_2020}. The second column from top to bottom shows the results of OARPD~\cite{Qiao_MTA_2024}, Yang~\etal{}~\cite{Yang_ICCV_2023}, and our method. Green and red rectangles indicate the ground-truth and estimated bounding boxes, respectively. All methods show detection results with confidence scores greater than 0.3.}
\label{fig-qualitative-result-0060_0390}
\end{figure*}

\begin{figure*}[t]
\centering
\includegraphics[width=0.75\hsize]{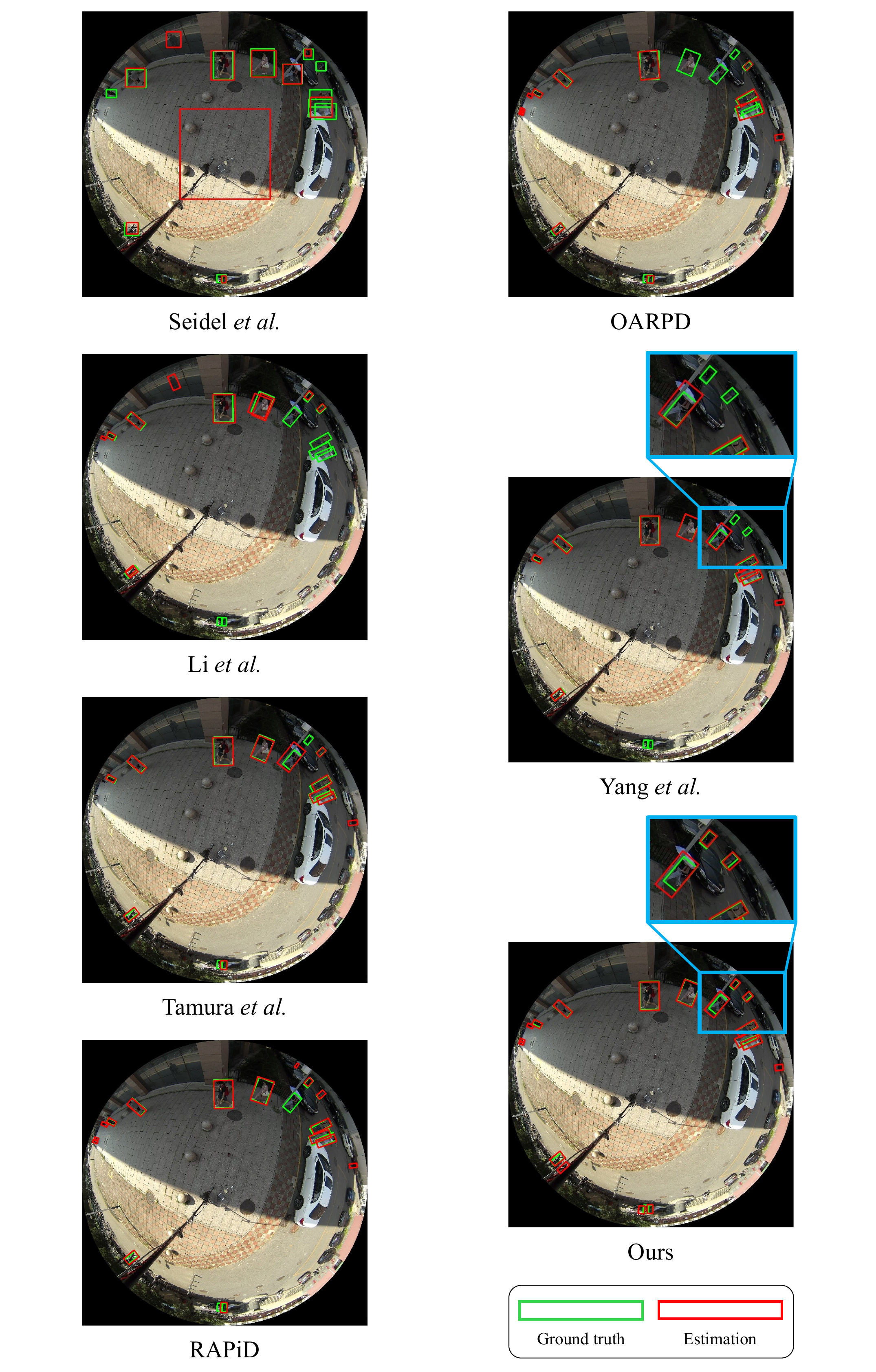}
\caption{Qualitative results achieved on the LOAF test set. The first column from top to bottom shows the results of methods of Seidel~\etal{}\cite{Seidel_VISAPP_2019}, Li~\etal{}~\cite{Li_AVSS_2019}, Tamura~\etal{}~\cite{Tamura_WACV_2019}, and RAPiD~\cite{Duan_CVPRW_2020}. The second column from top to bottom shows the results of OARPD~\cite{Qiao_MTA_2024}, Yang~\etal{}~\cite{Yang_ICCV_2023}, and our method. Green and red rectangles indicate the ground-truth and estimated bounding boxes, respectively. All methods show detection results with confidence scores greater than 0.3.}
\label{fig-qualitative-result-0064_01785}
\end{figure*}

To evaluate the detection accuracy qualitatively, we compared the bounding boxes on the LOAF test set. \fref{fig-qualitative-result}(a) shows our detection results in a panoramic image. Although persons at the top of the panoramic image appear small, our network can detect the bounding boxes accurately. We found that our method detected some small persons that were not included in the ground-truth annotations, as shown in the zoomed-in area of \fref{fig-qualitative-result}(a). Our network thus has the potential to detect more persons. Overall, our results were the closest to the ground-truth bounding boxes. The existing state-of-the-art method proposed by Yang~\etal{}~\cite{Yang_ICCV_2023} failed to detect some persons at the near image circles, as shown in \fref{fig-qualitative-result}(b). Moreover, \fref{fig-qualitative-result-0060_0390} and \fref{fig-qualitative-result-0064_01785} show qualitative results of conventional methods~\cite{{Duan_CVPRW_2020, Li_AVSS_2019, Qiao_MTA_2024, Seidel_VISAPP_2019, Tamura_WACV_2019, Yang_ICCV_2023}} and our method on the LOAF test set. Our method outperformed these conventional methods in terms of detection accuracy.

\begin{figure*}[t]
\centering
\includegraphics[width=0.75\hsize]{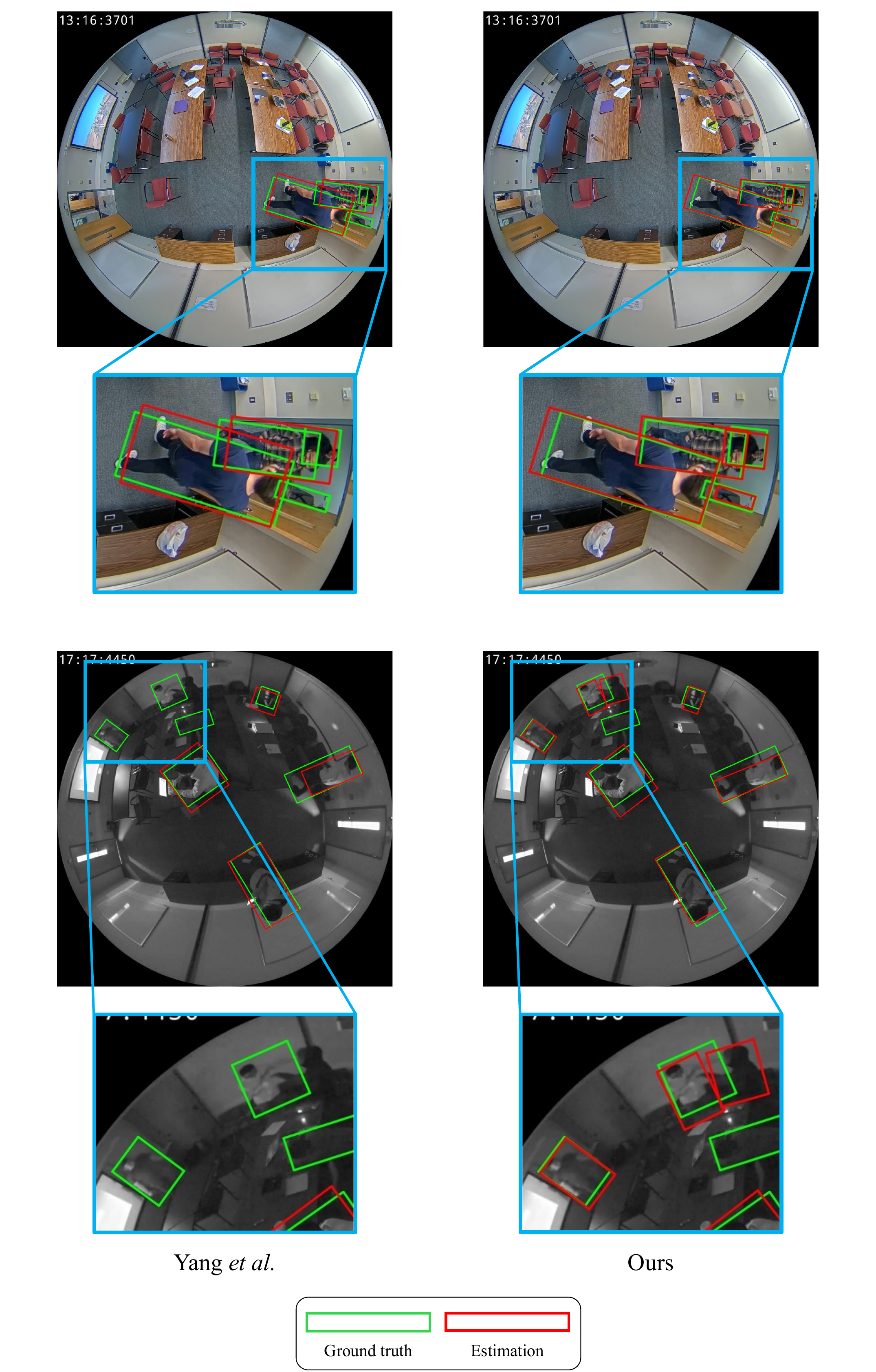}
\caption{Qualitative results achieved on the CEPDOF dataset. Green and red rectangles indicate the ground-truth and estimated bounding boxes, respectively. The detection results of Yang~\etal{}~\cite{Yang_ICCV_2023} on the left side and our method on the right side. All methods show detection results with confidence scores greater than 0.3.}
\label{fig-qualitative-result-cepdof}
\end{figure*}

\begin{figure*}[t]
\centering
\includegraphics[width=0.75\hsize]{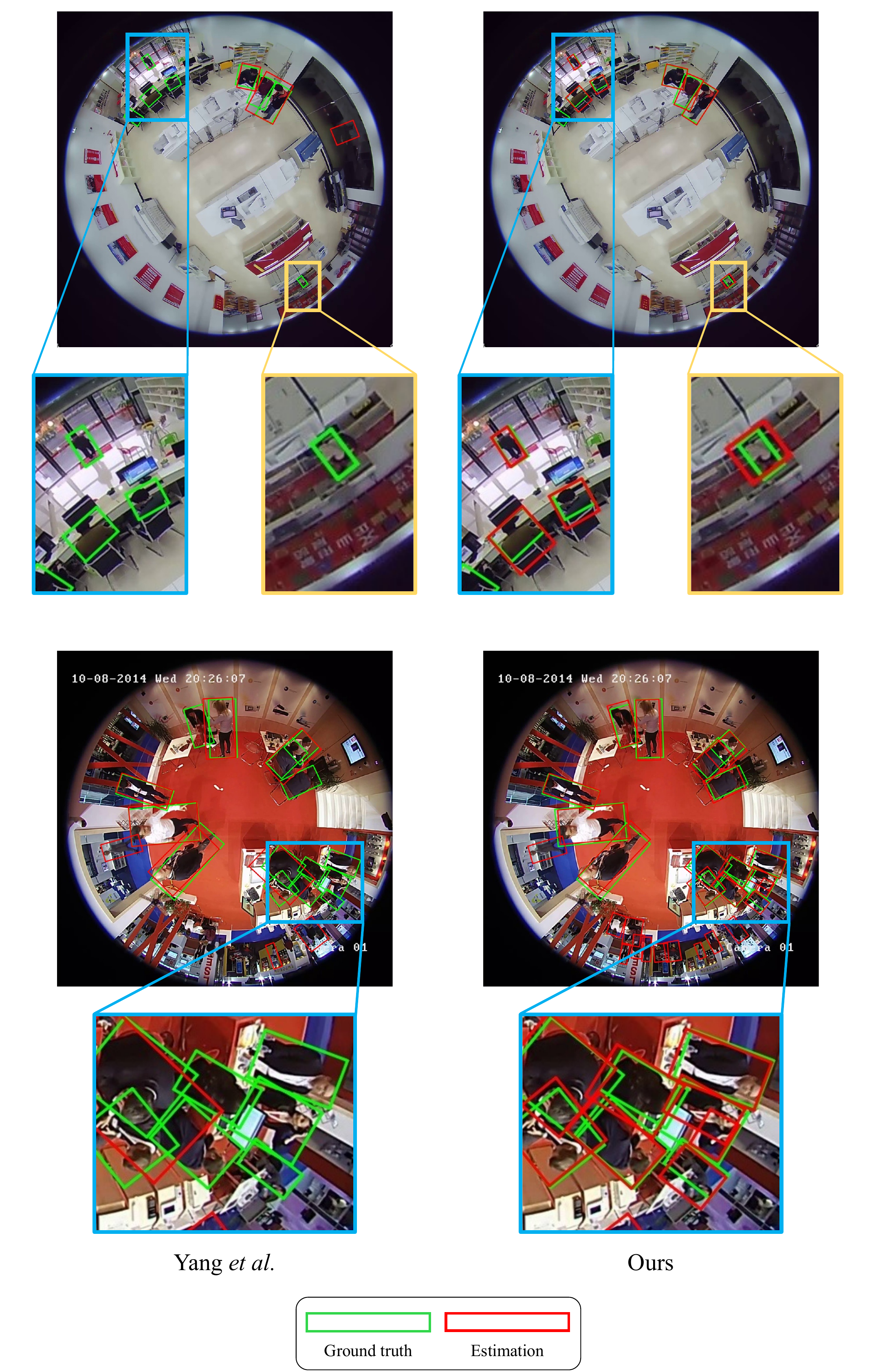}
\caption{Qualitative results achieved on the WEPDTOF dataset. Green and red rectangles indicate the ground-truth and estimated bounding boxes, respectively. The detection results of Yang~\etal{}~\cite{Yang_ICCV_2023} on the left side and our method on the right side. All methods show detection results with confidence scores greater than 0.3.}
\label{fig-qualitative-result-wepdtof}
\end{figure*}

\vspace{2mm}
We also evaluated our method on the CEPDOF and WEPDTOF datasets. As shown in \fref{fig-qualitative-result-cepdof}, overhead fisheye images in CEPDOF were captured in rooms, including scenes with low illuminance, while WEPDTOF contains fisheye images captured in various indoor environments, such as offices, stores, and exhibition rooms, as shown in \fref{fig-qualitative-result-wepdtof}. The qualitative results across both datasets show the limited performance of Yang~\etal{}'s method~\cite{Yang_ICCV_2023} in detecting occluded persons. This limitation arose from the differences between fisheye and panoramic image inputs. Person appearances in fisheye images exhibit greater variation than those in panoramic images. Therefore, the conventional method exhibited degraded detection performance when processing fisheye images. Specifically, fisheye images of occluded persons did not provide sufficient features for reliable detection. In contrast, our method using panoramic images demonstrated robust performance against occlusions. The detection results on CEPDOF and WEPDTOF showed that our method achieved more accurate detection than Yang~\etal{}'s method~\cite{Yang_ICCV_2023}. These results demonstrated that our method consistently outperformed the conventional methods in terms of detection accuracy across different datasets.

\clearpage

\clearpage

{
\small
\bibliographystyle{ieeenat_fullname}
\bibliography{egbib_long}

@string{ADICS = "{Proceedings of the International Conference on Advances in Data Engineering and Intelligent Computing Systems (ADICS)}"}

@string{AIVR = "{Proceedings of the IEEE International Conference on Artificial Intelligence and Virtual Reality (AIVR)}"}

@string{AVSS = "{Proceedings of the IEEE International Conference on Advanced Video and Signal Based Surveillance (AVSS)}"}

@string{CVPR = "{Proceedings of the IEEE Conference on Computer Vision and Pattern Recognition (CVPR)}"}

@string{CVPRCVF = "{Proceedings of the IEEE/CVF Conference on Computer Vision and Pattern Recognition (CVPR)}"}

@string{CVPRWCVF = "{Proceedings of the IEEE/CVF Conference on Computer Vision and Pattern Recognition Workshops (CVPRW)}"}

@string{ECCV = "{Proceedings of the European Conference on Computer Vision (ECCV)}"}

@string{ICCV = "{Proceedings of the IEEE International Conference on Computer Vision (ICCV)}"}

@string{ICCVCVF = "{Proceedings of the IEEE/CVF International Conference on Computer Vision (ICCV)}"}

@string{ICPR = "{Proceedings of the International Conference on Pattern Recognition (ICPR)}"}

@string{VISAPP = "{Proceedings of the International Conference on Computer Vision Theory and Applications (VISAPP)}"}

@string{WACV = "{Proceedings of the IEEE Winter Conference on Applications of Computer Vision (WACV)}"}

@string{WACVCVF = "{Proceedings of the IEEE/CVF Winter Conference on Applications of Computer Vision (WACV)}"}

@string{IJCV = "{International Journal of Computer Vision (IJCV)}"}

@string{JVCIR = "{Journal of Visual Communication and Image Representation (JVCIR)}"}

@string{MTA = "{Multimedia Tools and Applications (MTA)}"}

@string{PAMI = "{IEEE Transactions on Pattern Analysis and Machine Intelligence (PAMI)}"}

@string{SPIC = "{Signal Processing: Image Communication (SPIC)}"}

@string{TIP = "{IEEE Transactions on Image Processing (TIP)}"}

@string{ICLR = "{Proceedings of the International Conference on Learning Representations (ICLR)}"}

@string{ICRA = "{Proceedings of the IEEE International Conference on Robotics and Automation (ICRA)}"}

@string{IROS = "{Proceedings of the IEEE/RSJ International Conference on Intelligent Robots and Systems (IROS)}"}

@string{NIPS = "{Proceedings of the Advances in Neural Information Processing Systems (NeurIPS)}"}

@string{SCIENCE = "{Science}"}

@string{LMS = "{Proceedings of the London Mathematical Society}"}

@string{JEP = "{Journal of Experimental Psychology}"}

@INPROCEEDINGS{Wakai_CVPR_2024,
  author={Wakai, N. and Sato, S. and Ishii, Y. and Yamashita, T.},
  title={Deep Single Image Camera Calibration by Heatmap Regression to Recover Fisheye Images Under Manhattan World Assumption},
  booktitle=CVPRCVF,
  year={2024},
  pages={11884-11894},
  doi={10.1109/CVPR52733.2024.01129}
}

@INPROCEEDINGS{Wakai_ECCV_2022,
  author={Wakai, N. and Sato, S. and Ishii, Y. and Yamashita, T.},
  booktitle=ECCV,
  title={Rethinking Generic Camera Models for Deep Single Image Camera Calibration to Recover Rotation and Fisheye Distortion},
  year={2022},
  volume={13678},
  number={},
  pages={679-698},
  doi={10.1007/978-3-031-19797-0\_39}
}

@INPROCEEDINGS{Seidel_VISAPP_2019,
  author={Seidel, R. and Apitzsch, A. and Hirtz, G.},
  booktitle=VISAPP,
  title={Improved Person Detection on Omnidirectional Images with Non-maxima Suppression},
  year={2019},
  pages={},
  doi={}
}

@INPROCEEDINGS{Li_AVSS_2019,
  author={Li, S. and Ozan Tezcan, M. and Ishwar, P. and Konrad, J.},
  booktitle=AVSS,
  title={Supervised People Counting Using An Overhead Fisheye Camera},
  year={2019},
  pages={1-8},
  doi={10.1109/AVSS.2019.8909877}
}

@INPROCEEDINGS{Tamura_WACV_2019,
  author={Tamura, M. and Horiguchi, S. and Murakami, T.},
  booktitle=WACV, 
  title={Omnidirectional Pedestrian Detection by Rotation Invariant Training},
  year={2019},
  volume={},
  number={},
  pages={1989-1998},
  doi={10.1109/WACV.2019.00216}
}

@INPROCEEDINGS{Duan_CVPRW_2020,
  author={Duan, Z. and Ozan Tezcan, M. and Nakamura, H. and Ishwar, P. and Konrad, J.},
  booktitle=CVPRWCVF, 
  title={{RAPiD}: Rotation-Aware People Detection in Overhead Fisheye Images}, 
  year={2020},
  volume={},
  number={},
  pages={2700-2709},
  doi={10.1109/CVPRW50498.2020.00326}
}

@ARTICLE{Qiao_MTA_2024,
  author={Qiao, R. and Cai, C. and Meng, H and Wang, F. and Zhao, J.},
  journal=MTA,
  title={{OARPD}: Occlusion-aware rotated people detection in overhead fisheye images},
  year={2024},  
  volume={83},
  number={},
  pages={90375-90392},
  doi={10.1007/s11042-024-18852-2},
}

@INPROCEEDINGS{Yang_ICCV_2023,
  author={Yang, L. and Li, L. and Xin, X. and Sun, Y. and Song, Q. and Wang, W.},
  booktitle=ICCVCVF,
  title={Large-Scale Person Detection and Localization using Overhead Fisheye Cameras},
  year={2023},
  pages={19904-19914},
  doi={10.1109/ICCV51070.2023.01827}
}

@INPROCEEDINGS{Redmon_CVPR_2017,
  author={Redmon, J. and Farhadi, A.},
  title={{YOLO9000}: Better, Faster, Stronger}, 
  booktitle=CVPR, 
  year={2017},
  volume={},
  number={},
  pages={6517-6525},
  doi={10.1109/CVPR.2017.690}
}

@ARTICLE{Redmon_arXiv_2018,
  author={Redmon, J. and Farhadi, A.},
  journal={arXiv preprint arXiv:1804.02767},
  title={{YOLOv3}: An Incremental Improvement}, 
  year={2018},
  doi={10.48550/arXiv.1804.02767}
}

@INPROCEEDINGS{Varghese_ADICS_2024,
  author={Varghese, R. and M., S.},
  booktitle=ADICS, 
  title={{YOLOv8}: A Novel Object Detection Algorithm with Enhanced Performance and Robustness}, 
  year={2024},
  volume={},
  number={},
  pages={1-6},
  doi={10.1109/ADICS58448.2024.10533619}
}

@INPROCEEDINGS{Lin_CVPR_2017,
  author={Lin, T.-Y. and Doll\`{a}r, P. and Girshick, R. and He, K. and Hariharan, B. and Belongie, S.},
  booktitle=CVPR, 
  title={Feature Pyramid Networks for Object Detection}, 
  year={2017},
  volume={},
  number={},
  pages={936-944},
  doi={10.1109/CVPR.2017.106}
}

@INPROCEEDINGS{Liu_ICLR_2022,
  author={Liu, S. and Li, F. and Zhang, H. and Yang, X. and Qi, X. and Su, H. and Zhu, J and Zhang, L.},
  booktitle=ICLR,
  title={{DAB-DETR}: Dynamic Anchor Boxes are Better Queries for {DETR}},
  year={2022},
  pages={},
  doi={10.48550/arXiv.2201.12329}
}

@INPROCEEDINGS{Zhang_2025_CVPR,
  author={Zhang, C.-B. and Zhong, Y. and Han, K.},
  title= {{Mr. DETR}: Instructive Multi-Route Training for Detection Transformers},
  booktitle=CVPRCVF,
  year={2025},
  pages={9933-9943},
  doi={}
}

@INPROCEEDINGS{Huang_2025_CVPR,
  author={Huang, S. and Lu, Z. and Cun, X. and Yu, Y. and Zhou, X. and Shen, X.},
  title={{DEIM}: {DETR} with Improved Matching for Fast Convergence},
  booktitle=CVPRCVF,
  year={2025},
  pages={15162-15171}
}

@INPROCEEDINGS{Yuan_CVPR_2024,
  author={Yuan, T. and Zhang, X. and Liu, K. and Liu, B. and Chen, C. and Jin, J. and Jiao, Z.},
  booktitle=CVPRCVF, 
  title={Towards Surveillance Video-and-Language Understanding: New Dataset, Baselines, and Challenges}, 
  year={2024},
  volume={},
  number={},
  pages={22052-22061},
  doi={10.1109/CVPR52733.2024.02082}
}

@INPROCEEDINGS{Fu_CVPR_2021,
  author={Fu, Z. and Liu, Q. and Fu, Z. and Wang, Y.},
  booktitle=CVPRCVF,
  title={{STMTrack}: Template-free Visual Tracking with Space-time Memory Networks},
  year={2021},
  volume={},
  number={},
  pages={13774-13783},
  doi={10.1109/CVPR46437.2021.01356}
}

@INPROCEEDINGS{Xie_ICPR_2010,
  author={Xie, Y. and Lin, L. and Jia, Y.},
  booktitle=ICPR, 
  title={Tracking Objects with Adaptive Feature Patches for {PTZ} Camera Visual Surveillance}, 
  year={2010},
  volume={},
  number={},
  pages={1739-1742},
  doi={10.1109/ICPR.2010.430}
}

@ARTICLE{Lin_TIP_2012,
  author={Lin, L. and Lu, Y. and Pan, Y. and Chen, X.},
  journal=TIP, 
  title={Integrating Graph Partitioning and Matching for Trajectory Analysis in Video Surveillance}, 
  year={2012},
  volume={21},
  number={12},
  pages={4844-4857},
  doi={10.1109/TIP.2012.2211373}
}

@INPROCEEDINGS{Khan_CVPR_2023,
  author={Khan, A. H. and Nawaz, M. S. and Dengel, A.},
  booktitle=CVPRCVF, 
  title={Localized Semantic Feature Mixers for Efficient Pedestrian Detection in Autonomous Driving}, 
  year={2023},
  volume={},
  number={},
  pages={5476-5485},
  doi={10.1109/CVPR52729.2023.00530}
}

@INPROCEEDINGS{Dollar_CVPR_2009,
  author={Doll\`{a}r, P. and Wojek, C. and Schiele, B. and Perona, P.},
  booktitle=CVPR, 
  title={Pedestrian detection: A benchmark}, 
  year={2009},
  volume={},
  number={},
  pages={304-311},
  doi={10.1109/CVPR.2009.5206631}
}

@ARTICLE{Zhang_PAMI_2018,
  author={Zhang, S. and Benenson, R. and Omran, M. and Hosang, J. and Schiele, B.},
  journal=PAMI, 
  title={Towards Reaching Human Performance in Pedestrian Detection}, 
  year={2018},
  volume={40},
  number={4},
  pages={973-986},
  doi={10.1109/TPAMI.2017.2700460}
}

@INPROCEEDINGS{Yoshimi_IROS_2006,
  author={Yoshimi, T. and Nishiyama, M. and Sonoura, T. and Nakamoto, H. and Tokura, S. and Sato, H. and Ozaki, F. and Matsuhira, N. and Mizoguchi, H.},
  booktitle=IROS, 
  title={Development of a Person Following Robot with Vision Based Target Detection}, 
  year={2006},
  volume={},
  number={},
  pages={5286-5291},
  doi={10.1109/IROS.2006.282029}
}

@INPROCEEDINGS{Zhang_IROS_2024,
  author={Zhang, D. and Birner, L. and Pancheri, F. and Rehekampff, C. and Burschka, D. and Lueth, T. C.},
  booktitle=IROS, 
  title={A Hybrid Human Tracking System using {UWB} Sensors and Monocular Visual Data Fusion for Human Following Robots}, 
  year={2024},
  volume={},
  number={},
  pages={10878-10883},
  doi={10.1109/IROS58592.2024.10801756}
}

@INPROCEEDINGS{Lee_ICRA_2018,
  author={Lee, B.-J. and Choi, J. and Baek, C. and Zhang, B.-T.},
  booktitle=ICRA, 
  title={Robust Human Following by Deep Bayesian Trajectory Prediction for Home Service Robots}, 
  year={2018},
  volume={},
  number={},
  pages={7189-7195},
  doi={10.1109/ICRA.2018.8462969}
}

@ARTICLE{Torii_PAMI_2021,
  author={Torii, A and Taira, H and Sivic, J. and Pollefeys, M. and Okutomi, M. and Pajdla, T. and Sattler, T},
  journal=PAMI, 
  title={Are Large-Scale {3D} Models Really Necessary for Accurate Visual Localization?}, 
  year={2021},
  volume={43},
  number={3},
  pages={814-829},
  doi={10.1109/TPAMI.2019.2941876}
}

@INPROCEEDINGS{Balntas_ECCV_2018,
  author={Balntas, V. and Li, S. and Prisacariu, V.},
  booktitle=ECCV,
  title={{RelocNet}: Continuous metric learning relocalisation using neural nets},
  year={2018},
  volume={11218},
  pages={782-799},
  doi={10.1007/978-3-030-01264-9\_46}
}

@INPROCEEDINGS{YangL_ICCV_2019,
  author={Yang, L. and Bai, Z. and Tang, C. and Li, H. and Furukawa, Y. and Tan, P.},
  booktitle=ICCVCVF, 
  title={{SANet}: Scene Agnostic Network for Camera Localization}, 
  year={2019},
  volume={},
  number={},
  pages={42-51},
  doi={10.1109/ICCV.2019.00013}
}

@INPROCEEDINGS{Sarlin_CVPR_2021,
  author={Sarlin, P.-E. and Unagar, A. and Larsson, M. and Germain, H. and Toft, C. and Larsson, V. and Pollefeys, M. and Lepetit, V. and Hammarstrand, L. and Kahl, F. and Sattler, T.},
  booktitle=CVPRCVF, 
  title={Back to the Feature: Learning Robust Camera Localization from Pixels to Pose}, 
  year={2021},
  volume={},
  number={},
  pages={3246-3256},
  doi={10.1109/CVPR46437.2021.00326}
}

@INPROCEEDINGS{Hyeon_ICCV_2021,
  author={Hyeon, J. and Kim, J. and Doh, N.},
  booktitle=ICCVCVF, 
  title={Pose Correction for Highly Accurate Visual Localization in Large-scale Indoor Spaces}, 
  year={2021},
  volume={},
  number={},
  pages={15954-15963},
  doi={10.1109/ICCV48922.2021.01567}
}

@INPROCEEDINGS{Aiger_ICCV_2023,
  author={Aiger, D. and Araujo, A. and Lynen, S.},
  booktitle=ICCVCVF, 
  title={Yes, we {CANN}: Constrained Approximate Nearest Neighbors for local feature-based visual localization}, 
  year={2023},
  volume={},
  number={},
  pages={13293-13303},
  doi={10.1109/ICCV51070.2023.01227}
}

@INPROCEEDINGS{Sarlin_CVPR_2019,
  author={Sarlin, P.-E. and Cadena, C. and Siegwart, R. and Dymczyk, M.},
  booktitle=CVPRCVF, 
  title={From Coarse to Fine: Robust Hierarchical Localization at Large Scale}, 
  year={2019},
  volume={},
  number={},
  pages={12708-12717},
  doi={10.1109/CVPR.2019.01300}
}

@INPROCEEDINGS{Germain_CVPR_2021,
  author={Germain, H. and Lepetit, V. and Bourmaud, G.},
  booktitle=CVPRCVF, 
  title={Neural Reprojection Error: Merging Feature Learning and Camera Pose Estimation}, 
  year={2021},
  volume={},
  number={},
  pages={414-423},
  doi={10.1109/CVPR46437.2021.00048}
}

@INPROCEEDINGS{Kendall_ICCV_2015,
  author={Kendall, A. and Grimes, M. and Cipolla, R.},
  booktitle=ICCV,
  title={{PoseNet}: A Convolutional Network for Real-Time 6-{DOF} Camera Relocalization},
  year={2015},
  pages={2938-2946},
  doi={10.1109/ICCV.2015.336}
}

@INPROCEEDINGS{Walch_ICCV_2017,
  author={Walch, F. and Hazirbas, C. and Leal-Taixe, L. and Sattler, T. and Hilsenbeck, S. and Cremers, D.},
  booktitle=ICCV,
  title={Image-Based Localization Using {LSTMs} for Structured Feature Correlation},
  year={2017},
  pages={627-637},
  doi={10.1109/ICCV.2017.75},
}

@INPROCEEDINGS{Zong_ICCV_2023,
  author={Zong, Z. and Song, G. and Liu, Y.},
  booktitle=ICCVCVF, 
  title={{DETRs} with Collaborative Hybrid Assignments Training}, 
  year={2023},
  volume={},
  number={},
  pages={6725-6735},
  doi={10.1109/ICCV51070.2023.00621}
}

@INPROCEEDINGS{Ghiasi_CVPR_2021,
  author={Ghiasi, G. and Cui, Y. and Srinivas, A. and Qian, R. and Lin, T.-Y. and Cubuk, E. D. and Le, Q. V. and Zoph, B.},
  booktitle=CVPRCVF, 
  title={Simple Copy-Paste is a Strong Data Augmentation Method for Instance Segmentation}, 
  year={2021},
  volume={},
  number={},
  pages={2917-2927},
  doi={10.1109/CVPR46437.2021.00294}
}

@INPROCEEDINGS{Wang_CVPR_2023,
  author={Wang, W. and Dai, J. and Chen, Z. and Huang, Z. and Li, Z. and Zhu, X. and Hu, X. and Lu, T. and Lu, L. and Li, H. and Wang, X. and Qiao, Y.},
  booktitle=CVPRCVF, 
  title={{InternImage}: Exploring Large-Scale Vision Foundation Models with Deformable Convolutions}, 
  year={2023},
  volume={},
  number={},
  pages={14408-14419},
  doi={10.1109/CVPR52729.2023.01385}
}

@INPROCEEDINGS{Ren_NIPS_2015,
  author={Ren, S. and He, K. and Girshick, R. and Sun, J.},
  booktitle=NIPS,
  title={{Faster R-CNN}: {Towards} Real-Time Object Detection with Region Proposal Networks},
  year={2015},
  volume={28},
  number={},
  pages={},
  doi={10.48550/arXiv.1506.01497}
}

@INPROCEEDINGS{Liu_ECCV_2016,
  author={Liu, W. and Anguelov, D. and Erhan, D. and Szegedy, C. and Reed, S. and Fu, C.-Y. and Berg, A. C.},
  booktitle=ECCV,
  title={{SSD}: {Single} Shot MultiBox Detector},
  year={2016},
  volume={},
  number={},
  pages={},
  doi={}
}

@INPROCEEDINGS{Wang_NIPS_2024,
  author={Wang, A. and Chen, H. and Liu, L. and Chen, K. and Lin, Z. and Han, J. and Ding, G.},
  booktitle=NIPS,
  title={{YOLOv10}: real-time end-to-end object detection},
  year={2024},
  pages={},
  doi={10.48550/arXiv.2405.14458}
}

@INPROCEEDINGS{Tezcan_WACV_2022,
  author={Tezcan, M. O. and Duan, Z. and Cokbas, M. and Ishwar, P. and Konrad, J.},
  booktitle=WACVCVF, 
  title={{WEPDTOF}: A Dataset and Benchmark Algorithms for In-the-Wild People Detection and Tracking from Overhead Fisheye Cameras}, 
  year={2022},
  volume={},
  number={},
  pages={1381-1390},
  doi={10.1109/WACV51458.2022.00145}
}

@ARTICLE{Blanco_SPIC_2021,
  author={del-Blanco, C. R. and Carballeira, P. and Jaureguizar, F. and Garc{\'i}a, N.},
  journal=SPIC,
  title={Robust people indoor localization with omnidirectional cameras using a Grid of Spatial-Aware Classifiers},
  volume={93},
  pages={116135},
  year={2021},
  doi={10.1016/j.image.2021.116135}
}

@INPROCEEDINGS{Lin_ECCV_2014,
  author={Lin, T.-Y. and Maire, M. and Belongie, S. and Hays, J. and Perona, P. and Ramanan, D. and Doll{\'a}r, P. and Zitnick, C. L.},
  booktitle=ECCV,
  title={Microsoft {COCO}: Common Objects in Context},
  year={2014},
  volume={8693},
  pages={740-755},
  doi={10.1007/978-3-319-10602-1\_48}
}

@ARTICLE{Russakovsky_IJCV_2015,
 author={Russakovsky, O. and Deng, J. and Su, H. and Krause, J. and Satheesh, S. and Ma, S. and Huang, Z. and Karpathy, A. and Khosla, A. and Bernstein, M. and Berg, A. C. and Fei-Fei, L.},
 journal=IJCV,
 title={{ImageNet} Large Scale Visual Recognition Challenge},
 year={2015},
 volume={115},
 number={3},
 pages={211-252},
 doi={10.1007/s11263-015-0816-y}
}

@INPROCEEDINGS{Dosovitskiy_ICLR_2021,
  author={Dosovitskiy, A. and Beyer, L. and Kolesnikov, A. and Weissenborn, D. and Zhai, X. and Unterthiner, T. and Dehghani, M. and Minderer, M. and Heigold, G. and Gelly, S. and Uszkoreit, J. and Houlsby, N.},
  booktitle=ICLR,
  title={An Image is Worth 16x16 Words: Transformers for Image Recognition at Scale},
  year={2021},
  pages={},
  doi={10.48550/arXiv.2010.11929}
}

@INPROCEEDINGS{Liu_ICCV_2021,
  author={Liu, Z. and Lin, Y. and Cao, Y. and Hu, H. and Wei, Y. and Zhang, Z. and Lin, S. and Guo, B.},
  booktitle=ICCV,
  title={{Swin Transformer}: Hierarchical Vision Transformer using Shifted Windows},
  year={2021},
  pages={9992-10002},
  doi={10.1109/ICCV48922.2021.00986}
}

@INPROCEEDINGS{Hassani_CVPR_2023,
  author={Hassani, A. and Walton, S. and Li, J. and Li, S. and Shi, H.},
  booktitle=CVPRCVF,
  title={Neighborhood Attention Transformer},
  year={2023},
  volume={},
  number={},
  pages={6185-6194},
  doi={10.1109/CVPR52729.2023.00599}
}

@INPROCEEDINGS{Dong_CVPR_2022,
  author={Dong, X. and Bao, J. and Chen, D. and Zhang, W. and Yu, N. and Yuan, L. and Chen, D. and Guo, B.},
  booktitle=CVPRCVF, 
  title={{CSWin Transformer}: A General Vision Transformer Backbone with Cross-Shaped Windows}, 
  year={2022},
  volume={},
  number={},
  pages={12114-12124},
  doi={10.1109/CVPR52688.2022.01181}
}

@INPROCEEDINGS{Xia_CVPR_2022,
  author={Xia, Z. and Pan, X. and Song, S. and Li, L. E. and Huang, G.},
  booktitle=CVPRCVF, 
  title={Vision Transformer with Deformable Attention}, 
  year={2022},
  volume={},
  number={},
  pages={4784-4793},
  doi={10.1109/CVPR52688.2022.00475}
}

@INPROCEEDINGS{Rao_NIPS_2021,
  author={Rao, Y. and Zhao, W. and Liu, B. and Lu, J. and Zhou, J. and Hsieh, C.-J.},
  booktitle=NIPS, 
  title={{DynamicViT}: {Efficient} Vision Transformers with
Dynamic Token Sparsification}, 
  year={2021},
  volume={34},
  number={},
  pages={13937-13949},
  doi={10.48550/arXiv.2106.02034}
}

@INPROCEEDINGS{Zhu_CVPR_2023,
  author={Zhu, L. and Wang, X. and Ke, Z. and Zhang, W. and Lau, R.},
  booktitle=CVPRCVF, 
  title={{BiFormer}: Vision Transformer with Bi-Level Routing Attention}, 
  year={2023},
  volume={},
  number={},
  pages={10323-10333},
  doi={10.1109/CVPR52729.2023.00995}
}

@ARTICLE{Wang_PAMI_2024,
  author={Wang, W. and Chen, W. and Qiu, Q. and Chen, L. and Wu, B. and Lin, B. and He, X. and Liu, W.},
  authorfull={Wang, Wenxiao and Chen, Wei and Qiu, Qibo and Chen, Long and Wu, Boxi and Lin, Binbin and He, Xiaofei and Liu, Wei},
  journal=PAMI, 
  title={{CrossFormer++}: A Versatile Vision Transformer Hinging on Cross-Scale Attention}, 
  year={2024},
  volume={46},
  number={5},
  pages={3123-3136},
  doi={10.1109/TPAMI.2023.3341806}
}

@INPROCEEDINGS{Tu_ECCV_2022,
  author={Tu, Z. and Talebi, H. and Zhang, H. and Yang, F. and Milanfar, P. and Bovik, A. and Li, Y.},
  booktitle=ECCV,
  title={{MaxViT}: Multi-Axis Vision Transformer},
  year={2022},
  volume={13684},
  pages={459–479},
  doi={10.1007/978-3-031-20053-3\_27}
}

@INPROCEEDINGS{Athwale_ICCV_2023,
  author={Athwale, A. and Afrasiyabi, A. and Lag\"{u}e, J. and Shili, I. and Ahmad, O. and Lalonde, J.-F.},
  booktitle=ICCVCVF, 
  title={{DarSwin}: Distortion Aware Radial Swin Transformer}, 
  year={2023},
  volume={},
  number={},
  pages={5906-5915},
  doi={10.1109/ICCV51070.2023.00545}
}

@INPROCEEDINGS{Ren_ICCV_2023,
  author={Ren, S. and Yang, X. and Liu, S. and Wang, X.},
  booktitle=ICCVCVF,
  title={{SG-Former}: Self-guided Transformer with Evolving Token Reallocation}, 
  year={2023},
  volume={},
  number={},
  pages={5980-5991},
  doi={10.1109/ICCV51070.2023.00552}
}

@INPROCEEDINGS{Plaut_CVPRW_2021,
  author={Plaut, E. and Ben Yaacov, E. and El Shlomo, B.},
  booktitle=CVPRWCVF, 
  title={{3D} Object Detection from a Single Fisheye Image Without a Single Fisheye Training Image}, 
  year={2021},
  volume={},
  number={},
  pages={3654-3662},
  doi={10.1109/CVPRW53098.2021.00405}
}

@INPROCEEDINGS{Wang_CVPR_2021,
  author = {Wang, F.-E. and Yeh, Y.-H. and Sun, M. and Chiu, W.-C. and Tsai, Y.-H.},
  booktitle=CVPRCVF,
  title={{LED2-Net}: Monocular $360^{\circ}$ Layout Estimation via Differentiable Depth Rendering},
  year={2021},
  volume={},
  number={},
  pages={12951-12960},
  doi={10.1109/CVPR46437.2021.01276}
}

@INPROCEEDINGS{Krizhevsky_NIPS_2012,
  author={Krizhevsky, A. and Sutskever, I. and Hinton, G. E.},
  booktitle=NIPS,
  title={{ImageNet} Classification with Deep Convolutional Neural Networks},
  volume={25},
  year={2012},
  doi={10.1145/3065386}
}

@INPROCEEDINGS{Vaswani_NIPS_2017,
  author={Vaswani, A. and Shazeer, N. and Parmar, N. and Uszkoreit, J. and Jones, L. and Gomez, A. N. and Kaiser, \L{}. and Polosukhin, I.},
  booktitle=NIPS,
  title={Attention Is All You Need},
  year={2017},
  pages={6000-6010},
  doi={10.48550/arXiv.1706.03762}
}

@INPROCEEDINGS{Lopez_CVPR_2019,
 author={M. {L\'{o}pez}-Antequera and R. {Mar\'{i}} and P. {Gargallo} and Y. {Kuang} and J. {Gonzalez-Jimenez} and G. {Haro}},
 booktitle=CVPRCVF,
 title={Deep Single Image Camera Calibration With Radial Distortion},
 year={2019},
 volume={},
 number={},
 pages={11809-11817},
 doi={10.1109/CVPR.2019.01209}
}

@INPROCEEDINGS{Dal_ECCV_2024,
  author={Dal Cin, A. P. and Azzoni, F. and Boracchi, G. and Magri, L.},
  booktitle=ECCV,
  title={Revisiting Calibration of Wide-Angle Radially Symmetric Cameras},
  year={2024},
  volume={15094},
  pages={214–230},
  doi={10.1007/978-3-031-72764-1\_13}
}

@ARTICLE{Zhu_arXiv_2018,
  author={Zhu, P. and Wen, L. and Bian, X. and Ling, H. and Hu, Q.},
  journal={arXiv preprint arXiv:1804.07437},
  title={Vision meets drones: A challenge}, 
  year={2018},
  doi={10.48550/arXiv.1804.07437}
}

@INPROCEEDINGS{Singh_NIPS_2018,
  author={Singh, B. and Najibi, M. and Davis, L. S.},
  booktitle=NIPS, 
  title={{SNIPER}: Efficient Multi-Scale Training}, 
  year={2018},
  volume={},
  number={},
  pages={9310-9320},
  doi={10.48550/arXiv.1805.09300}
}

@INPROCEEDINGS{Singh_CVPR_2018,
  author={Singh, B. and Davis, L. S.},
  booktitle=CVPRCVF, 
  title={An Analysis of Scale Invariance in Object Detection - {SNIP}}, 
  year={2018},
  volume={},
  number={},
  pages={3578-3587},
  doi={10.1109/CVPR.2018.00377}
}

@INPROCEEDINGS{Gong_WACV_2021,
  author={Gong, Y. and Yu, X. and Ding, Y. and Peng, X. and Zhao, J. and Han, Z.},
  booktitle=WACV,
  title={Effective Fusion Factor in FPN for Tiny Object Detection},
  year={2021},
  volume={},
  pages={1159-1167},
  doi={10.1109/WACV48630.2021.00120},
}

@INPROCEEDINGS{Yu_WACV_2020,
  author={Yu, X. and Gong, Y. and Jiang, N. and Ye, Q. and Han, Z.},
  booktitle=WACV, 
  title={Scale Match for Tiny Person Detection}, 
  year={2020},
  volume={},
  number={},
  pages={1246-1254},
  doi={10.1109/WACV45572.2020.9093394}
}

@INPROCEEDINGS{Unel_CVPRW_2019,
  author={\"{U}nel, F. \"{O}. and \"{O}zkalayci, B. O. and \c{C}i\v{g}la, C.},
  booktitle=CVPRWCVF, 
  title={The Power of Tiling for Small Object Detection}, 
  year={2019},
  volume={},
  number={},
  pages={582-591},
  doi={10.1109/CVPRW.2019.00084}
}

@ARTICLE{Liu_TIP_2025,
  author={Liu, K. and Fu, Z. and Jin, S. and Chen, Z. and Zhou, F. and Jiang, R. and Chen, Y. and Ye, J.},
  journal=TIP, 
  title={{ESOD}: Efficient Small Object Detection on High-Resolution Images}, 
  year={2025},
  volume={34},
  number={},
  pages={183-195},
  doi={10.1109/TIP.2024.3501853}
}

@ARTICLE{Wang_PAMI_2021,
  author={Wang, J. and Sun, K. and Cheng, T. and Jiang, B. and Deng, C. and Zhao, Y. and Liu, D. and Mu, Y. and Tan, M. and Wang, X. and Liu, W. and Xiao, B.},
  journal=PAMI, 
  title={Deep High-Resolution Representation Learning for Visual Recognition}, 
  year={2021},
  volume={43},
  number={10},
  pages={3349-3364},
  doi={10.1109/TPAMI.2020.2983686}
}

@INPROCEEDINGS{Ji_ICPR_2021,
  author={Ji, H. and Gao, Z. and Liu, X. and Zhang, Y. and Mei, T.},
  booktitle=ICPR, 
  title={Small Object Detection Leveraging on Simultaneous Super-resolution}, 
  year={2021},
  volume={},
  number={},
  pages={803-810},
  doi={10.1109/ICPR48806.2021.9413058}
}

@INPROCEEDINGS{Shermeyer_CVPRW_2019,
  author={Shermeyer, J. and Van Etten, A.},
  booktitle=CVPRWCVF, 
  title={The Effects of Super-Resolution on Object Detection Performance in Satellite Imagery}, 
  year={2019},
  volume={},
  number={},
  pages={1432-1441},
  doi={10.1109/CVPRW.2019.00184}
}

@INPROCEEDINGS{Sun_2025_CVPR,
  author={Sun, H. and Wang, R. and Li, Y. and Yang, L. and Lin, S. and Cao, X. and Zhang, B.},
  title={{SET}: Spectral Enhancement for Tiny Object Detection},
  booktitle=CVPRCVF,
  year={2025},
  pages={4713-4723}
}

@ARTICLE{DelBlanco_SPIC_2021,
  author={Del-Blanco, C. R. and Carballeira, P. and Jaureguizar, F. and Garc\'{i}a, N.},
  authorfull={Del-Blanco, Carlos R. and Carballeira, Pablo and Jaureguizar, Fernando and Garc\'{i}a, Narciso},
  journal=SPIC,
  title={Robust people indoor localization with omnidirectional cameras using a {Grid of Spatial-Aware Classifiers}},
  year={2021},
  volume={93},
  number={},
  pages={116135},
  doi={10.1016/j.image.2021.116135},
}

@ARTICLE{Zhu_JVCIR_2019,
  author={Zhu, J. and Zhu, J. and Wan, X. and Wu, C. and Xu, C.},
  authorfull={Zhu, Jun and Zhu, Jiangcheng and Wan, Xudong and Wu, Chao and Xu, Chao},
  journal=JVCIR,
  title={Object detection and localization in {3D} environment by fusing raw fisheye image and attitude data},
  year={2019},
  volume={59},
  number={},
  pages={128-139},
  doi={10.1016/j.jvcir.2019.01.005},
}

@INPROCEEDINGS{Sharma_AIVR_2019,
  author={Sharma, A. and Ventura, J.},
  booktitle=AIVR, 
  title={Unsupervised Learning of Depth and Ego-Motion From Cylindrical Panoramic Video}, 
  year={2019},
  volume={},
  number={},
  pages={58-587},
  doi={10.1109/AIVR46125.2019.00018}
}

@ARTICLE{Mandelbrot_Science_1967,
  author={Mandelbrot, B.},
  journal=SCIENCE,
  title={How Long Is the Coast of Britain? Statistical Self-Similarity and Fractional Dimension},
  year={1967},
  volume={156},
  number={3775},
  pages={636-638},
  doi={10.1126/science.156.3775.636},
}

@ARTICLE{Smith_LMS_1874,
  author={Smith, H. J. S.},
  journal=LMS,
  title={On the integration of discontinuous functions},
  year={1874},
  volume={s1-6},
  number={1},
  pages={140-153},
  doi={10.1112/plms/s1-6.1.140},
}

@ARTICLE{Gibson_JEP_1959,
  author={Gibson, E. J. and Gibson, J. J. and Smith, O. W. and Flock, H.},
  journal=JEP,
  title={Motion parallax as a determinant of perceived depth},
  year={1959},
  volume={58},
  number={1},
  pages={40-51},
  doi={10.1037/h0043883},
}

@BOOK{Hartley_CUP_2004,
  author={Hartley, R. and Zisserman, A.},
  title={Multiple View Geometry in Computer Vision},
  edition={Second},
  year={2004},
  publisher={Cambridge University Press},
  doi={10.1017/CBO9780511811685}
}

@INPROCEEDINGS{Loshchilov_ICLR_2019,
  author={Loshchilov, I. and Hutter, F.},
  booktitle=ICLR,
  title={Decoupled Weight Decay Regularization},
  year={2019},
  pages={},
  doi={10.48550/arXiv.1711.05101}
}

@INPROCEEDINGS{Paszke_NIPS_2019,
  author={Paszke, A. and Gross, S. and Massa, F. and Lerer, A. and Bradbury, J. and Chanan, G. and Killeen, T. and Lin, Z. and Gimelshein, N. and Antiga, L. and Desmaison, A. and K\"{o}pf, A. and Yang, E. and DeVito, Z. and Raison, M. and Tejani, A. and Chilamkurthy, S. and Steiner, B. and Fang, L. and Bai, J. and Chintala, S.},
  title={{PyTorch}: An Imperative Style, High-Performance Deep Learning Library},
  booktitle=NIPS,
  year={2019},
  pages={8024-8035},
  doi={10.48550/arXiv.1912.01703}
}

@MISC{MMCV_MISC_2018,
  title={{MMCV: OpenMMLab} Computer Vision Foundation},
  author={{MMCV} Contributors},
  howpublished={\url{https://github.com/open-mmlab/mmcv}},
  year={2018}
}
}

\end{document}